\documentclass[Afour,sageh,times]{sagej}

\usepackage{moreverb,url}
\usepackage[colorlinks, linkcolor=black,bookmarksopen,bookmarksnumbered,citecolor=black,urlcolor=black]{hyperref}

\usepackage{flushend}
\usepackage{graphicx}
\usepackage{algorithm}
\usepackage{algorithmicx}
\usepackage[noend]{algpseudocode}
\usepackage{color}
\usepackage[dvipsnames]{xcolor}
\usepackage{amsmath,amssymb,amsfonts,dsfont}
\usepackage{bm}
\usepackage{setspace}
\usepackage{multirow}
\usepackage{mathtools}
\usepackage{fancyhdr}
\usepackage{authblk}
\usepackage{url}
\usepackage{tikz}

\usepackage{wrapfig}    
\usepackage{mymath,renew}      
\usepackage{pifont}     
\usepackage[font=footnotesize]{caption}
\usepackage{balance}

\usepackage[
  separate-uncertainty = true,
  multi-part-units = repeat
]{siunitx}
\usepackage{threeparttable}
\usetikzlibrary{decorations.pathreplacing}

\usepackage{renew}      

\newcommand\BibTeX{{\rmfamily B\kern-.05em \textsc{i\kern-.025em b}\kern-.08em
T\kern-.1667em\lower.7ex\hbox{E}\kern-.125emX}}

\newcommand{\lossDetB}{\ensuremath{{\mathcal L}_{\textrm{det}}}}
\newcommand{\lossClass}{\ensuremath{{\mathcal L}_{\textrm{cls}}}}
\newcommand{\lossBBox}{\ensuremath{{\mathcal L}_{\textrm{bb}}}}
\newcommand{\lossAttrib}{\ensuremath{{\mathcal L}_{\textrm{att}}}}
\newcommand{\lossAfford}{\ensuremath{{\mathcal L}_{\textrm{aff}}}}
\newcommand{\lossRPN}{\ensuremath{{\mathcal L}_{\textrm{rpn}}}}
\newcommand{\lossTotal}{\ensuremath{{\mathcal L}_{\textrm{tot}}}}
\newcommand{\zr}{\;--\,}
\newcommand{\balg}{\textit}
\newcommand{\algName}{{AffContext}}
\newcommand{\algNameKL}{{AffContext$_{KL-att}$}}
\renewcommand{\seclabel}{\label}

\usepackage[switch]{lineno} 
\usepackage{amsmath}   
\begin{document}
\runninghead{Chu, Xu, Tang and Vela}

\title{Recognizing object affordances to support scene reasoning for
manipulation tasks}

\author{Fu-Jen Chu\affilnum{1}, Ruinian Xu\affilnum{1}, Chao Tang\affilnum{1} and Patricio A. Vela\affilnum{1}}

\affiliation{
\affilnum{1}School of Electrical and Computer Engineering, 
    Georgia Institute of Technology, GA, USA.
}

\corrauth{Fu-Jen Chu, 
Intelligent Vision and Automation Laboratory (IVALab),
School of Electrical and Computer Engineering,
Georgia Institute of Technology,
North Ave NW,
Atlanta, 
GA 30332.
}

\email{fujenchu@gatech.edu}

\begin{abstract}
Affordance information about a scene provides important clues as to what
actions may be executed in pursuit of meeting a specified goal state.
Thus, integrating affordance-based reasoning into symbolic action
plannning pipelines would enhance the flexibility of robot manipulation.
Unfortunately, the top performing affordance recognition methods use
object category priors to boost the accuracy of affordance detection and
segmentation. Object priors limit generalization to unknown object
categories.  This paper describes an affordance recognition pipeline
based on a category-agnostic region proposal network for proposing
instance regions of an image across categories. 
To guide affordance learning in the absence of category priors, the
training process includes the auxiliary task of explicitly inferencing
existing affordances within a proposal. Secondly, a self-attention
mechanism trained to interpret each proposal learns to capture rich
contextual dependencies through the region. Visual benchmarking shows
that the trained network, called \balg{\algName}, reduces the
performance gap between object-agnostic and object-informed affordance
recognition.
\balg{\algName} is linked to the Planning Domain Definition Language
(PDDL) with an augmented state keeper for action planning across
temporally spaced goal-oriented tasks.
Manipulation experiments show that \balg{\algName} can successfully
parse scene content to seed a symbolic planner problem specification,
whose execution completes the target task. 
Additionally,  task-oriented grasping for cutting and pounding actions
demonstrate the exploitation of multiple affordances for a given object
to complete specified tasks. 
\end{abstract}

\keywords{Recognition, Manipulation Planning, Grasping, AI Reasoning Methods}
\maketitle

\section{Introduction}
Identifying the functionalities, or \textit{affordances}, of object
parts aids task completion by informing robot manipulators on how to
use or interact with an object.
Adult humans possess rich prior knowledge for recognizing
affordances of object parts. 
The affordance knowledge supports identification of potential
interactions with nearby objects, and contributes to planning
manipulation sequences with these objects towards achievement of
a defined task.
Endowing a robot with the same capabilities is crucial for assistive
robots operating in human environments on a daily basis. 
%
%

Affordance detection of object parts in images is frequently cast as a
segmentation problem
\citep{myers2015affordance, srikantha2016weakly,nguyen2016detecting,%
roy2016multi, nguyen2017object, do2018affordancenet}. 
Object parts sharing the same functionality are segmented and grouped
at the pixel level, then assigned the corresponding affordance label.
This problem formulation permits the use of state-of-the-art semantic
segmentation architectures based on convolutional neural networks
(CNNs) derived by the computer vision community 
\citep{krizhevsky2012imagenet}.
However, affordance identification differs from conventional
semantic segmentation based on visual cues or physical properties, and
need not require assigning labels to all image pixels. 
Understanding functionalities of an object part requires learning the
concept of potential interactions with--or use by--humans.
%
State-of-the-art affordance detection architectures 
\citep{nguyen2017object, do2018affordancenet} improve segmentation
performance by jointly optimizing for object detection and affordance
prediction. 
The object prior and instance features inferred from the detection process
improve pixel-wise affordance map predictions.
Given that per-object bounding box and per-pixel label annotations are
labor-intensive, 
existing affordance datasets contain few object categories
when compared to commonly seen classification datasets
\citep{krizhevsky2009learning, deng2009imagenet, openimages}
in the vision community. 
Consequently, a limited amount of object categories can be learned with
these datasets, while the open-world involves more diverse categories. 
In effect, the specificity of object priors limits generalizability.

%
Generalizing learnt affordances across novel object categories is
essential for open-world affordance recognition and would more 
fully utilize affordance annotated datasets. 
Decoupling the object category prior from the instance feature space
performing affordance prediction removes the specificity. It also
undermines affordance estimation since affordances are typically
associated to object parts. 
Replacing the object category detection process with an
\textit{objectness} detection branch parallel to the affordance
segmentation bridges the gap for novel categories while maintaining the
benefits of utilizing local features for predicting object part
affordances, but incurs a performance drop \citep{chu2019toward}.
When the internal feature space does not exhibit the richness or
performance needed, one means to compensate for information loss 
in the network is to incorporate attention mechanisms 
\citep{vaswani2017attention}, in particular attention for image-level
segmentation tasks \citep{yuan2018ocnet, zhang2018context, fu2019dual}.

\begin{figure}[t]
  \centering
  \begin{tikzpicture}[inner sep=0pt,outer sep=0pt]
  \node (SF) at (0in,0in)
    {\includegraphics[height=1.5in,clip=true,trim=0.0in 0.0in 0.0in 0.0in]{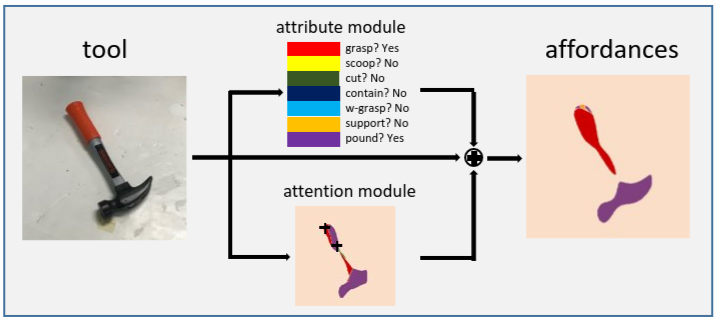}};
  \end{tikzpicture}
  \caption{
  Affordance detection framework with the proposed attribute and
  attention modules. The attribute and attention modules improve
  pixel-wise prediction of object part affordances. 
  The attribute module (upper branch) predicts affordances for a
  region of interest as shareable attributes across categories. 
  As an auxiliary task, it guides the local region feature learning.
  The attention module (bottom branch) learns dependencies across
  pixels. For example, the two \textit{plus marks} on the hammer's
  handle with high correlation should have the same predicted affordance
  labels.\label{fig_concept}}
\end{figure}

Our prior work \citep{chu2019toward} described a category-agnostic
affordance detection architecture with a parallel \textit{objectness}
detection branch, which could be interpreted as an instance-based
attention module. 
With instance features agnostic to object label, the segmentation branch
training process implicitly learns features attuned to pixel-level
affordances for each object proposal.  The generalized approach
sacrifices performance and exhibits an affordance segmentation score
drop of 20\% relative to object-aware approaches.
To mitigate the performance drop, we explore the addition of
an intra-region processing module to learn long-range or non-local
affordance relationships across instance contexts. 
In affordance prediction networks, this type of
attention module 
%
supports high-level understanding by capturing correlations between
affordances and other appearance cues (such as object form). 
Internalizing these correlations improves segmentation accuracy.
Furthermore, affordances are shared across the object parts of different
objects with similar functions (and possibly similar forms).
Thus seen or known affordances may be present in unseen or novel object
categories observed in the open-world.   
To learn visual cues indicative of function, affordances are
also used as object attributes shared across object categories in the
\textit{objectness} branch.
%
%
In lieu of strong object priors from an object detector, affordance
attribute recognition serves to guide the object instance regions to
learn shareable features across novel object categories. 
Fig. \ref{fig_concept} illustrates the general concept of the proposed
affordance detection modules with attention and attribute learning.
For a given object region proposal, the \textit{objectness} detection
branch leverages affordance attributes to enhance detection via
affordance context. Likewise, the segmentation branch leverages
attention to promote improved awareness of affordances within
an image region proposal. The approach, called \balg{\algName}, reduces the
existing performance gap and supports specification based manipulation
using a basic manipulation grammar with affordance-informed predicates.
%
%
\begin{figure}[t]
  \centering
  \vspace*{0.08in}
  \begin{tikzpicture}[inner sep=0pt,outer sep=0pt]
  \node (SF) at (0in,0in)
    {\includegraphics[height=3.5in,clip=true,trim=0.0in 0.0in 0.0in 0.0in]{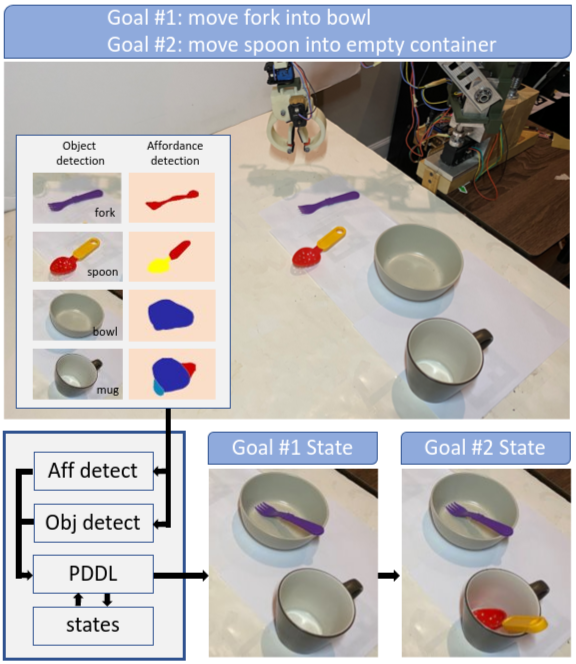}};

  \node[anchor=north east,xshift=0pt,yshift=-3pt] at (SF.north west) {(a)};
  \node[anchor=north east,xshift=0pt,yshift=-53pt] at (SF.north west) {(b)}; 
  \node[anchor=north east,xshift=0pt,yshift=-165pt] at (SF.north west) {(c)}; 
    
  \end{tikzpicture}
  \caption{
  Illustration of the agnostic affordance detection framework
  with PDDL for goal-directed physical robotic manipulations.
  \textbf{(a)} The overall goal is to first move the fork into the bowl, and then move the spoon into the mug. \textit{Goal \#1}               explicitly specifies the object to grasp (fork), and the object to contain (bowl). \textit{Goal \#2}, however, requires knowledge from previously achieved goal state; 
  \textbf{(b)} The proposed category-agnostic affordance detector predicts possible actions to be performed on object parts. Together with a pre-trained object detector, both objects and affordances in the robot's view are identified; 
  \textbf{(c)} Given a goal state, detected objects and affordances form the initial state for PDDL to plan an action sequence for execution. Given a second goal state, the previous goal state contributes to current initial state for PDDL to inference and plan.         
    \label{fig_concept_PDDL}
    }
\end{figure}
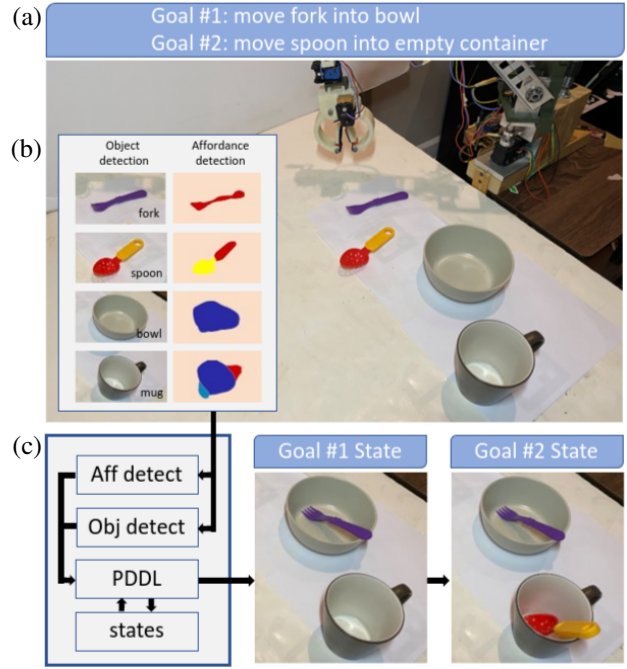

For planning with affordances, our earlier work \citep{chu2019toward}
defined a symbolic reasoning framework with Planning Domain Definition
Language (PDDL) \citep{mcdermott1998pddl} for achieving simple goal-oriented tasks using the detected affordances. This framework is
more extensively tested and extended to exploit detected object-action 
pairs for goal-oriented tasks.  It adds a separate object detection
pipeline applicable to a richer set of object categories than exist for
existing affordance datasets.
As shown in Fig. \ref{fig_concept_PDDL}, the object category-agnostic
affordance detector coordinates can coordinate a (separately) pretrained
object detector for flexible object selection during visual processing.
In practice some tasks may involve multiple sub-goals, where visual
information fails to seed the symbolic world state due to occlusion;
keeping track of status of objects in a scene is required. 
We add state memory from previous manipulation actions and their terminal
states to preserve knowledge of the world state even if an action
resulted in occlusions (and therefore in detection failures for the
occluded objects).  The overall framework translates affordances to
planned action primitives for achieving multiple goal states in any
order.
\subsubsection{Contribution.}
This work is grounded on the premise that endowing an agent with
affordance knowledge about its local environment from visual sensors
provides potential action opportunities using nearby objects, which
then contributes to hypothesizing action plans to achieve a goal.
For robot agents to identify affordances and recognize potential
interactions in an object diverse open-world, they need to understand
affordances as functional properties that abstract beyond a specific set
of objects, but that are shared by many objects with the similar form.
As a step in this direction, this manuscript
describes the \balg{\algName} deep network  with the object
category-agnostic affordance segmentation.  It closes the performance
gap between object-aware and object-agnostic affordance segmentation
approaches.
\balg{\algName} predicts object-action pairs to infer action
opportunities on each object part across unseen categories. 
The obtained affordance knowledge converted into symbolic state
information supports symbolic reasoning and planning with action
primitives using affordance-based predicate specifications.
Specifically, a system for goal-oriented task execution is presented
to generate manipulation sequences by translating affordances to action
primitives given a goal. 
%
Several robotic manipulation experiments ranging from simple movements,
to task-oriented grasping, to goal-oriented tasks demonstrate that the
approach translates to actual manipulation for an embodied robotic arm.
Performance of \balg{\algName} is close to that of a state-of-the-art
object-aware affordance recognition approach but also applies to unseen
objects with the same performance.
The success rate is around 96\% for affordance recognition and passive
exploitation (e.g., \textit{grasp}, \textit{contain}, \textit{support}), 
and the task completion rate is 88\% for active affordance exploitation
using objects as tools (e.g., \textit{cut}, \textit{pound}, \textit{scoop}).
The results demonstrate that endowing a robot agent with affordance
recognition capabilities provides it with a means to convert the
recognized affordances (and objects) into executable action primitives
based on simple specifications. 

\section{Related Work}
%
%


Due to its primacy and role in robotic manipulation, the most commonly
studied affordance has been the \textit{grasp} affordance 
\citep{shimoga1996robot, bicchi2000robotic, bohg2014data}.
Machine learning is increasingly being used for grasping based on the
ability to learn to grasp objects \citep{saxena2008robotic} for which
generative or model-based methods would be difficult.
Recent literature in this vein includes using a cascaded network \citep{lenz2015deep}
to encode grasp features, and using deep networks to
learn graspable areas \citep{ngiam2011multimodal} and 
graspable regions \citep{redmon2015real, GuEtAl_ICRA2017, chu2018deep}
.
Alternatively, mapping from vision input to manipulation actions can be
learned through a large collection of physical demonstrations or
interactions, with \cite{levine2016learning} initially taking a
supervised approach later extended to reinforcement learning
\citep{kalashnikov2018qt}.
Robotics research on general affordances, beyond grasping,
studies the potential interactions of robots with their surrounding
objects and/or environments 
\citep{ugur2015bottom,dehban2016denoising,nguyen2016preparatory}.
Detecting the affordances of object parts in images is cast as a
pixel-wise labelling problem and exploits computer vision approaches to
segmentation.  
Object parts sharing the same functionality are grouped with the
corresponding ground truth affordance label. 
Detecting affordance involves identifying applicable actions on
object parts and segmenting the parts. 

\subsubsection{Affordance Recognition.}
%
Geometric cues with manually designed features were utilized for
pixel-wise affordance prediction in \cite{myers2015affordance}. 
The need for feature engineering was shifted to affordance-specific
feature learning using an encoder-decoder convolutional neural network
(CNN) architecture \citep{nguyen2016detecting}. 
Later, the image-based approach was extended to a
two-stage method \citep{nguyen2017object} where regional features
were obtained by applying object detection for object proposals on
an image. Assisted by object detection, the network predicted object
affordance on selected proposals from an entire input image.
Building on \cite{he2017mask}, the two-stage method was then
improved by jointly optimizing object detection and affordance
segmentation end-to-end \citep{do2018affordancenet}.
The object category priors from the object detection branch enhance
the pixel-level affordance estimates. 
%
%
%
Despite the state-of-the-art performance, annotation for detection
and segmentation are labor-intensive. 
To reduce annotation demands in supervised learning, weakly 
supervised approaches instead relied on sparse key point
annotations \citep{sawatzky2017weakly, sawatzky2017adaptive}.
To fully avoid the requirement of costly annotation process,
unsupervised learning on self-generated synthetic data was applied
followed by domain adaptation techniques \citep{chu2019learning}.
%
Since our work focuses on zero-order affordance \citep{aldoma2012supervised}
where affordances are functionalities found on objects and are irrespective
of current states in the world, self-supervised approaches
\citep{florence2018dense,zeng2018learning} lie outside of the scope of the
investigation. 

Optimizing detection and segmentation boosts performance through the
availability of object and location priors \citep{do2018affordancenet}
but limits the transfer of learned affordance labels to unseen
categories. 
To decouple the object category prior from the instance feature space, 
one solution is to detect \textit{objectness} of a region proposal
instead of predicting the object class \citep{chu2019toward}, thereby enabling
category-agnostic affordance segmentation on (object) instance features.
Isolating object priors from regional features permits inter-object
category prediction. However, the loss of object label priors or
constraints on processing requires alternative mechanisms to induce
region-driven learning or spatial aggregation.

\subsubsection{Attention and Auxiliary Tasks in Deep Networks.}

One means to induce contextual aggregation is to rely on object
attributes.
Attributes, as human describable properties, are known to assist vision
tasks, such as 
face detection \citep{kumar2009attribute, kumar2011describable}, 
object classification \citep{kumar2011describable, duan2012discovering}, 
activity recognition \citep{cheng2013nuactiv},
and fashion prediction \citep{liu2016deepfashion}. 
Affordances should also serve as shareable features with semantic meaning
whose use could benefit feature learning for object instances.
Attribute categories replace the discarded object categories during
training to guide feature learning, with the aim of improving
generalizability across novel object categories with recognized
affordances.
We extend \cite{chu2019toward} by employing attribute
prediction to guide instance feature learning while removing object 
category supervision. 
The attributes include affordance as a semantic label and
additional self-annotated visual attributes.

Recent studies on semantic segmentation enhance contextual aggregation by 
\textit{atrous} spatial pyramid pooling and dilated convolutions 
\citep{chen2018deeplab,chen2017rethinking}, 
merging information at various scales \citep{zhao2017pyramid}, and
fusing semantic features between levels \citep{ding2018context}. 
To model long-range pixel or channel dependencies in a feature map, 
attention modules \citep{vaswani2017attention} are applied in 
semantic segmentation for learning global dependencies 
\citep{yuan2018ocnet,zhang2018context,fu2019dual}. 
Region-based contextual aggregation in semantic segmentation or
object-based affordance segmentation remains unexplored.
Following prior works \citep{yuan2018ocnet, fu2019dual}, we propose to
incorporate a self-attention mechanism to model long-range
intra-regional dependencies. 
The proposed architecture adopts an attention mechanism in the
affordance branch and operates on object-based feature maps. 
Decision dependencies draw from local regions instead of the whole image.
The intent behind the architecture is to guide the instance features
with attribute learning, and model the dependencies within the instance
feature map for affordance prediction. The improved affordance knowledge
serves to aid real-world robotic manipulations.

\subsubsection{Symbolic Reasoning for Robotics.}
%
This subsection provides a overview of task-level symbolic reasoning in
robotics, with reasoning for manipulation provided in the next
subsection. Once scene elements have been recognized, conversion of the
information to support task or planning objectives is essential for
robot agents to achieve specified goals. Establishing how to
achieve the goal state from the known world state involves symbolic
reasoning based methods that identify atomic actions to take in order to meet the
goal specification \citep{ghallab2004automated}. 
Symbolic planning requires planning-domain languages such as 
STRIPS \citep{fikes1971strips}, 
ADL \citep{pednault1994adl},
PDDL \citep{mcdermott1998pddl}, or 
HAL \citep{marthi2007angelic}. 
Task-level planning with symbolic reasoning through atomic actions
decouples the steps from their continuous time implemenation, which can fail
to address barriers to execution.
%
Combining symbolic planning with motion planning
\citep{erdem2011combining, srivastava2014combined} or through a
hierarchical structure bridging the high-level specification and the
low-level actions \citep{kaelbling2011hierarchical, de2013towards}
addresses the potential feasibility gap when considering discrete actions
only. Geometric planning information establishes when pre- or
post-action conditions can truly be met based on known scene structure.
The motion planner may incorporate pyhsics engines to understand
potential dynamic interactions between the robot and objects in the
world and how they may open blocked paths
\citep{stilman2005navigation, levihn2014using, akbari2015ontological}.
Likewise, differential or dynamic constraints on robot movement may need
to be integrated into the motion planning module
\citep{de2013towards, cambon2004robot, plaku2010sampling}.
Ultimately, conversion of sensor data and the goal state to a fully
specified planning objective requires explicit knowledge of the object
and action content of the world. To be sufficiently general, this
knowledge is usually encoded within a compatible data structure whose
contents are derived from an ontology
\citep{tenorth2009knowrob,chen2013ontology}. What is needed is a
mechanism to connect visually sensor data to the corresponding elements
in the ontology and their associated action opportunities. It often
relies on visual recognition algorithms paired to items in the
ontology
\citep{tenorth2009knowrob,chen2013ontology,pandey2012towards,gravot2006cooking}
or recognition with pairwise geometric associations
\citep{zeng2018semantic}. 

This work focuses on the perception to specification component. 
Low-level motion planning is simplified through action primitives with
scripted plans, whose boundary conditions are determined from visual
information.  Object layout will be sufficiently sparse that motion
planning terminal conditions are always met and motion planning can be
presumed to succeed.  We tackle simple, task-level reasoning for
manipulation based on symbolic object and action knowledge extracted
from visual recognition algorithms.

\subsubsection{Symbolic Reasoning for Manipulation.}
With regards to scene understanding, robotic manipulation research that
focuses on the ontology and the underlying reasoning framework employs
(or assumes the existence of) tagged objects or recognition algorithms
tuned to the objects used
\citep{tenorth2009knowrob,chen2013ontology,pandey2012towards,zeng2018semantic}. 
The setup simplifies the perception problem. 
A less relaxed scene simplification is to have it consist of planar
objects, boxlike objects, or uniquely colored objects 
\citep{ugur2015bottom, gaschler2015extending, migimatsu2020object,
suarez2018interleaving,winkler2012knowledge}. 
Alternatively, it may rely on simulation
\citep{stilman2005navigation, akbari2016task, gravot2006cooking,suarez2018interleaving}
since object properties in simulation are precisely known and action
execution can be guaranteed (through reduced fidelity to real-world
dynamics).
These implementations create an experiment to reality gap in the
robotic system that prevents more general use of the underlying
ontological framework or manipulation strategies.  
However, they also permit deeper investigation into joint symbolic and
motion planning strategies that incorporate more advanced elements, 
such as probabilistic planning, optimality, linguistic imprecision, and
action execution failure detection
\citep{toussaint2010integrated,winkler2012knowledge}. 
%

With regards to the action specifications, this symbolic knowledge can
be formed by interactive experiences with the environment through
manipulation exploration \citep{ugur2015bottom}. Achieving the same
outcomes without relying on simple, coded scenarios in order to work for
more realistic and less constrained visual content requires connecting
to advances in computer vision \citep{yang2015robot}.  However, there is
a tension between 
inferring models from annotated data or experience 
versus 
using pre-specified models \citep{chen2013ontology}.
Following \cite{chen2013ontology}, we first aim to use a pre-specified
model in this paper before considering the augmentation of the action
specifications through demonstration or additional learning (which will
be left to a future investigation). Emphasis will be on using known
affordance categories, as perceived from visual sensors, to seed
symbolic planning specifications for manipulation.

%
%
%
%
On the perception to specification front, manipulation with symbolic
reasoning involving visual processing mainly deals with object poses
and inter-object spatial relationships \citep{sui2017goal, zeng2018semantic}.
Task specifications involve rearranging a scene to match a given target
arrangement. Doing so requires estimating scene graphs for the 6D poses
and inter-object relations of objects from RGB-D input imagery 
\citep{liu2020table}.
The scene graph describes object configurations using a tree structure,
which can then translate to specifications for goal-directed robot
manipulation plans in low clutter settings
\citep{sui2017goal,zeng2018semantic}.
Both the initial and goal states are given as  RGB-D images to be
converted into estimated poses and relations described within the
axiomatic scene graph, whose transcription to PDDL supports symbolic
planning.
Other forms of symbolic knowledge besides spatial properties can also be
used to define manipulation planning objectives, with affordances and
their intra-object interactions being one important class of symbolic
knowledge. Affordances can support general purpose vision-based
manipulation planning by providing awareness of potentially executable
atomic actions for a given scene.
Using predefined object-task affordances, high-level symbolic knowledge
is shown to improve prediction of grasping location for specific tasks
with low-level visual shape features \citep{antanas2019semantic}, which 
tackles the linkage between symbolic planning and geometric planning.
Scene understanding involving affordance recognition on realistic
objects is less explored.

Commonly
seen affordances are considered in \cite{antanas2019semantic}. However,
the affordances are predefined instead of predicted from visual context.
We aim to bridge the research gap of predicting affordance information
from visual input to support reasoning via symbolic planning for solving
goal-oriented manipulation tasks. 
Specifically, we consider predicates modeled from object part
functionality (and object states) as opposed to object spatial
relations. The available action states are inferred from the scene
through affordance estimates (and object recognition in some cases).  We
show that visual input can be converted into symbolic knowledge for use
by a PDDL planning solver to generate a sequence of executable atomic
actions whose execution achieves the desired goal state.

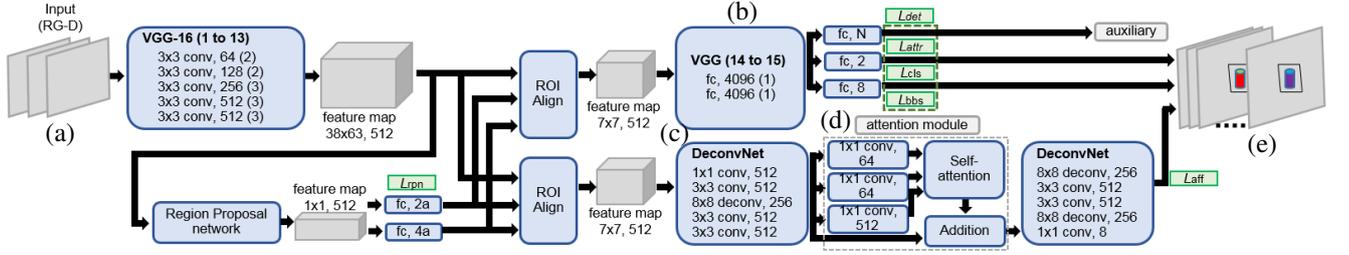
\begin{figure*}[t]
  \centering
  \scalebox{1.00}{
  \begin{tikzpicture}[inner sep=0pt,outer sep=0pt]
    \node (SF) at (0in,0in)
      {{\includegraphics[width=1\textwidth]{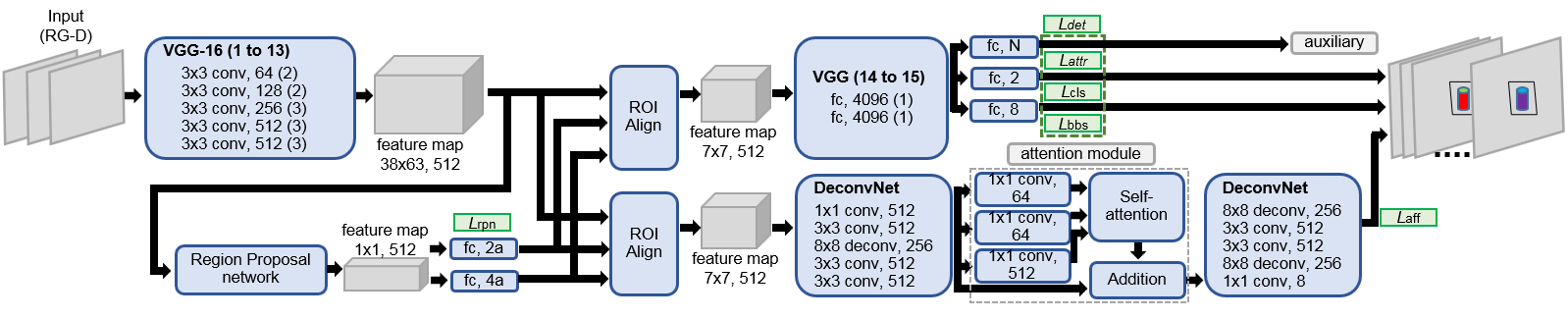}}};

    \node[anchor=north west,xshift=15pt,yshift=-48pt] at (SF.north west) {(a)};
    \node[anchor=north west,xshift=270pt,yshift=-2pt] at (SF.north west) {(b)};
    \node[anchor=north west,xshift=245pt,yshift=-48pt] at (SF.north west) {(c)};
    \node[anchor=north west,xshift=305pt,yshift=-43pt] at (SF.north west) {(d)};
    \node[anchor=north west,xshift=465pt,yshift=-52pt] at (SF.north west) {(e)};
    \end{tikzpicture}}
    \caption{Network structure of the proposed detector with self-attention and attribute learning. The network predicts affordances of object parts
	  for each object in the view. Blue blocks indicate network layers
	  and gray blocks indicate images and feature maps. 
    \textbf{(a)} RG-D images are input of the network; 
	\textbf{(b)} Category-agnostic proposals with \textit{objectness} are forced to predict attributes during training as an auxiliary task; 
    \textbf{(c)} Deconvolutional layers lead to a fine-grained 
	   feature map for learning long-range dependencies; 
    \textbf{(d)} Self-attention mechanism operation is incorporated in affordance branch on the intermediate feature ($30\times30\times512$); 
    \textbf{(e)} The final output includes bounding boxes and multiple layers 
	  indicating confidences for affordances on a single pixel.  
    \label{fig_structure}}
\end{figure*}

\section{Approach \seclabel{Approach}} 
There are two main objectives that aim to be linked to achieve
specification-based manipulation supported by affordance reasoning.
Given a corresponding pair of color and depth images, the first
objective is to identify pixel-wise affordances of object parts for seen
and unseen object categories. 
%
%
The second objective is to connect the affordance reasoning module with
a symbolic planning system so that the recognized affordances can seed
the initial state of the symbolic planner. The provision of a specified
goal state would then generate a sequence of manipulation actions that
change the initial state of the world to meet the goal state.
This section covers the design of the affordance recognition CNN
and its merger with a PDDL planner.

\subsection{Basic CNN Design}
%
The general structure of the deep network design for jointly predicting
object regions and their affordance segmentations across novel
categories is depicted in Fig.~\ref{fig_structure}. The network, called
\balg{\algName}, adopts a two-stage architecture 
\citep{he2017mask, do2018affordancenet} 
with VGG-16 \citep{simonyan2014very} as a backbone. 
It builds on earlier work \citep{chu2019toward} with a similar pipeline,
but includes modifications intended to improve affordance segmentation.
The first stage generates region proposals whose feature descriptors are
then input to separate detection and segmentation branches for
predicting object regions and affordance maps, respectively. 
More specifically, the shared feature map ($38\times63\times512$
feature) from the intermediate convolutional layers (layer 13 of VGG-16)
are sent to the \textit{Region Proposal network} for region proposals;
the two ROI align layers \citep{he2017mask} feed the collected instances
to the task branches.
After the first stage, two processing branches follow.
To support generalized learning for the segmentation branch (bottom) to
novel categories, the detection branch (top) performs binary
classification to separate foreground objects from the background. 
The segmentation branch takes in category-agnostic object regions for
predicting the affordance map within each region. 
To address the contextual dependencies and the non-local feature
learning within a region proposal, we introduce two improvements,
attention and attributes, described next.  They enhance the associations
among local features and guide feature learning to be object aware but
not object specific. 

\subsection{Region-based Self-Attention} 

The goal of object part affordance segmentation is to group pixels
sharing the same functionality and to assign them the correct
affordance labels. 
In urban street semantic segmentation \citep{Cordts2016Cityscapes, Richter_2016_ECCV}, 
the entire scene usually corresponds to large subset of possible
ground truth labels. In contrast, affordance segmentation assigns
labels to object regions only, with the assigned labels being a small
subset relative to the set of known affordance labels.
%
The semantic context of the image region (e.g., a cup) narrows the set of
relevant affordances (e.g., grasp, contain) and thus reduces the search
space \citep{zhang2018context}.  Since object category semantic priors
require object-specific detection modules that are not part of the
network design, alternative means to inform the segmentation are
necessary. Given that form and function are often correlated, the
missing object prior information could be recovered by having the
pixel-wise decisions depend on non-local pixel features that may already
encode for visual structure (form) related to affordances (function).
Doing so compensates for the small receptive field of convolutional
operations and provides more global context for local pixel-wise
decisions.

\begin{figure}
\includegraphics[width=8.5cm]{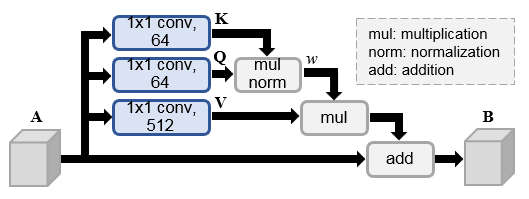}
\caption{Details of the attention module for regional features in affodance branch. $\mathbf{K}$, $\mathbf{Q}$ and $\mathbf{V}$ refer to key, query and value, respectively} \label{attentionModule}
\end{figure} 

To incorporate non-local contextual information, the network
architecture and learning process explicitly creates associations
between local features of pixels within a region proposal. 
%
A self-attention mechanism, as depicted in Fig.  \ref{attentionModule}, 
on the segmentation branch adapts long-range contextual information.
Given an instance feature map 
$\mathbf{A} \in \mathbb{R}^{c\times u\times v}$, 
a triplet of 
\textit{key} $\mathbf{K} \in \mathbb{R}^{c\times u\times v}$, 
\textit{query} $\mathbf{Q} \in \mathbb{R}^{c\times u\times v}$ and 
\textit{value} $\mathbf{V} \in \mathbb{R}^{c\times u\times v}$ 
feature maps are predicted.
The contextual relationship $\textit{w}_{ji}$ uses a spatial attention
module for features at pixel positions $j$ and $i$ within the instance
feature map:
\begin{equation} \label{contextual}
  \textit{w}_{ji} = \frac{1}{Z_j} \exp(Q^\top_i K_j),
\end{equation} 
where $\textit{w}_{ji}$ indicates the degree that the $\textit{i}^{th}$
representation feature influences the $\textit{j}^{th}$ feature. $Z_j$ is the normalization term:
\begin{equation} \label{normalization}
  Z_j = \sum_{i=1}^{u\times v} \exp(Q^\top_i K_j).
\end{equation} 
%
%
To aggregate the predicted correlation between features in different
positions within a region proposal, the feature map $\mathbf{V}$ is
associated with contextual relationship $\textit{w}_{ji}$ and learns a
residual function with the original feature map  $\mathbf{A}$ to provide
the final output $\mathbf{B} \in \mathbb{R}^{c\times u\times v}$. For
features at pixel position $j$:
\begin{equation} \label{residual}
B_j = \alpha \sum _{i=1}^{u \times v} (\textit{w}_{ji}V_i) + A_j
\end{equation} 
where $\alpha$ is a learnt scale parameter that
influences the weighting between the original feature vector $A_j$
and the attentional feature vector.

\subsection{Affordance as Auxiliary Task and Attribute} 

Having an object detection branch with binary classification removes
contextual information provided by the object category.
While it does provide category-agnostic processing, it does not
leverage potential information that may be transferable to unseen
object instances. 
Thus, in addition to predicting \textit{objectness}
through object detection, the detection pathway is augmented with
a multi-class classification module. 
The module works as an auxiliary task to explicitly predict existing affordances in a candidate region. 
The network is trained to
predict existing affordances across categories, which transfer to
unseen categories during deployment.
%
%
%
%
%
Predicting affordances at the pixel-level and at the region level are
related processes. Though a segmentation model may internally learn to predict
possible affordances within in a region, explicitly guiding the learning
process using an auxiliary task may allow two tasks to positively
influence each other 
\citep{jaderberg2016reinforcement, li2017deep, zhang2015learning}. 
%

%
Since affordance labels apply to objects outside of the training set,
predicting existing affordances for a region proposal can be treated as
predicting shareable attributes across object categories.
Attribute learning enhances object classification when suitable
attributes are available
\citep{farhadi2009describing, li2010objects, sun2013attribute}. 
In the auxiliary module, affordances are directly
treated as attributes to enhance the learning of visual representation.
Based on the potentially similar visual properties of objects with
similar functions, affordances may be recognizable attributes of unseen
objects based on their visual appearance. If this relationship can be
captured by incorporating attribute estimation into the detection
branch, then it will enhance learning visual representations related to
affordance.
%
%

For clarity, below we refer to affordances in the auxiliary module as
\textit{attributes} (or \textit{affordance attributes} in long form),
and reserve the simple term \textit{affordance} for the pixel-wise
segmentation case.
%
To guide the feature learning of each region proposal, a sub-branch
parallel to \textit{objectness} detection and bounding box regression
is augmented to perform attribute prediction with $N$ outputs, where $N$ is
the number of affordance attributes defined across categories.
Each attribute output is a binary classification predicting whether a
specific attribute (affordance) is found in the region proposal, based
on the instance feature shared with \textit{objectness}
detection and bounding box regression.

%
The \textit{objectness} branch identifies foreground from background
and hence the class number is $\mathbf{C} = 2$.
Let 
$\rho \in \Real^1$ denote the probability of an instance being foreground, 
$\beta \in \Real^4$ denote the corresponding bounding box, and 
$\alpha \in \Real^N$ denote the corresponding probabilities of attributes 
within the instance region. 
Define the loss function of the complete detection branch (\lossDetB) to be:
\begin{multline} \label{detection} 
  \lossDetB (\{(\rho, \beta, \alpha)\}_{c=0}^{\mathbf{C}-1}) 
    = \sum_c \lossClass(\rho)  \\
      + \lambda_1 \sum_c \delta_{c,1} \lossBBox(\beta_c, \beta_c^\ast) 
      + \frac{\lambda_2}{N} \sum_c \delta_{c,1} 
            \sum_i^N \lossAttrib (\alpha_i).
\end{multline}
where $\lossClass$ denotes the 
cross entropy loss for \textit{objectness} classification (cls), 
$\lossBBox$ denotes the $l_1$ loss for bounding box (bb) regression with 
$\beta_c^\ast$ the ground truth annotation, and
$\lossAttrib$ denotes the binary cross entropy loss for each attribute. 
The scalars $\lambda_1$ and $\lambda_2$ are optimization weight
factors, and $\delta_{\cdot, \cdot}$ is the Kronecker delta
function.

\subsection{Regional Attention and Attribute Embedding}
The regional attention and attribute learning modules are added to
the final network.  
The attribute learning parallel to the foreground detection works as
an auxiliary task to guide feature learning during training; it is
discarded during inference. 
The attention module in the segmentation branch learns a representation
of a region proposal that gathers contextual information; it is applied
during inference. 
To have a higher resolution affordance mask and to learn long-range
dependencies, deconvolutional layers initially upsample the feature
map of the segmentation branch (bottom in Fig.~\ref{fig_structure}).
Attention is applied after the first deconvolutional operation on
the $30\times30$ feature map, followed by two deconvolutional operations
for the final $244\times244$ affordance map.
To compute the loss for the affordance, let 
$q(j, a)$ denote the predicted affordance mask on a RoI-based feature map, 
where $j \in RoI$ is the $\textit{j}^{th}$ pixel in a region proposal,
and $a \in A$ is the $\textit{a}^{th}$ affordance at the pixel. 
The affordance loss $\lossAfford$ is defined as multinomial cross entropy loss:
\begin{equation} \label{L_aff}
  \lossAfford =  -\sum_{j \in RoI} \frac{1}{|RoI|} \sum_{a \in A} 
                                 Y(j,a) \log ( q(j,a) ) 
\end{equation}
where $|RoI|$ is the total area of the region of interest, $Y$ is the ground truth of the corresponding affordance mask with $A$ channels. 

The overall network inherits Faster-RCNN \citep{ren2015faster} on a 
VGG16 \citep{simonyan2014very} backbone with modified detection and
segmentation branches while keeping the region proposal network (RPN)
intact.
Let $\lossRPN$ denote the RPN loss from the original network, the
loss for the entire network is:
\begin{equation} \label{L_overall}
  \lossTotal = \lossDetB + \lossRPN + \lossAfford.
\end{equation}

\subsection{Planning with PDDL \subseclabel{statePDDL}}
Goal-oriented manipulation may require an agent to achieve a goal state
via a sequence of atomic actions, with each action involving an intermediate
state change of the objects in the environment and/or the manipulator.  
As a standard and widely used planning language, Planning Domain Definition Languange (PDDL) is adopted to encode and generate the desired sequence of actions.
The PDDL takes a \textbf{domain} definition and a \textbf{problem} description. The \textbf{domain} pre-defines a list of \textbf{predicates} and corresponding effects, while the \textbf{problem} includes the initial state and goal state descriptions.
To utilize the PDDL, the initial state required by the PDDL algorithm is
acquird via predictions for objectness and the predicted objects'
corresponding affordances in the image. Together with the goal state,
the planned sequence is solved by Fast Downward \citep{helmert2006fast}
as a list of executable atomic actions for a robotic manipulator.
%

To handle scenarios where determining an object category is required,
incorporating a general pre-trained object detector enables selecting
from multiple tools with similar functionality in the view, improving
the flexibility of use cases.    
%
To further handle scenarios where an intermediate goal exists, a
\textit{state keeper} is added to the PDDL to preserve the terminal
state from the previous planning session. 
Such a mechanism is essential since both detectors (\textit{objectness}
and the additional \textit{object-specific}) tend to miss an object when
it is inside of another object.  
Persistence of planned state information aids multi-step or sequential
planning as depicted in Fig. \ref{fig_concept_PDDL}, where the
sequentially applied goal states benefit from state persistence.
Incorporating \textit{object detector} and \textit{state keeper} with
the PDDL allows goal-oriented manipulation with tool selections, as
well as reusing states for consecutive goals. Goals such as
``\textit{place an object into a specific container and then place the
second object into any empty container}'' become possible.   

\section{Vision Evaluation \seclabel{EvalVis}}

This section describes the training process and benchmarking results of
the affordance prediction network \balg{\algName}. Relative to
the complete perceive, plan, act processing pipeline of a complete
implementation on a robotic arm, it focuses on the perception component
and tests performance as a visual processing algorithm absent the
embodied manipulation components. The training dataset is described, as
well as the training method, followed by the baseline approaches and
benchmarking results.

\subsection{UMD Dataset}
The UMD dataset \citep{myers2015affordance} covers 17 object categories with 7
affordances. The objects are from kitchen, workshop, and garden settings.
The dataset contains 28k+ RGB-D images captured by a Kinect sensor
with the object on a rotating table for data collection.  
The annotated segmentations provide affordance label ground truth 
for each object part.
The additional ground truth of object bounding boxes is obtained by
filtering out the background table from the foreground objects, and
establishing a tight rectangular bounding box. 
The UMD dataset has \textit{image split} and \textit{category split}
benchmarking approaches. Evaluation uses the \textit{category split}
benchmark, which tests unseen categories. 

\subsection{Data Preprocessing and Training}
The proposed approach reuses the weights of VGG-16
\citep{simonyan2014very} pre-trained on ImageNet \citep{deng2009imagenet} for initialization. 
The layers for attribute prediction and for the affordance branch,
including the attention module, are trained from scratch.
To incorporate RGB-D images for geometric information with
pre-trained weights, the blue channel is substituted with the depth
channel \citep{redmon2015real, chu2018deep}. Ideally,
any channel can be replaced with the depth channel.
The value of depth channel is normalized to the range $[0, 255]$ with
$144$ as the mean value. Missing value or $NaN$ in depth channel is
filled with $0$.
The whole network is trained end-to-end for $5$ epochs. The training
starts with initial learning rate $r = 0.001$ which is divided by
$10$ for every $2$ epochs. Training time is around $3$ days with
a single Nvidia GTX 1080 Ti.

\subsection{Baseline Methods}

In addition to including published outcomes for the UMD Benchmarking
Dataset, several sensible approaches serve as baselines. They arise from
partial or alternative implementation of \balg{\algName} algorithm, based on
the earlier work \citep{chu2019toward}.
%
The first \textit{baseline} approach, labelled \balg{Obj-wise}, is a
deep network structure trained with \textit{objectness} detection only,
i.e., without regional attention, and without attribute embedding.  
It is a modified version of \textit{AffNet} \citep{do2018affordancenet}
with object category labels unused during training on the
\textit{category-split} data.  
The network only outputs the primary affordance.
A second \textit{baseline} approach, denoted \balg{KLdiv}
\citep{chu2019toward}, uses the \balg{Obj-wise} network and replaces
the cross-entropy loss, \lossAfford, with KL-divergence for ranked
affordance outputs.
Specifically, since each object part is assigned from one to three ranked
affordances by human annotators, \balg{KLdiv} treats the ranking of
affordances as a distribution and learns to predict ranking during
inference time. The ranking output permits secondary affordance of an
object part. 
A third \textit{baseline}, denoted \balg{Multi}, also employs the
\balg{Obj-wise} network. To permit prediction of multiple affordances
on the same object part, \balg{Multi} simply replicates the
segmentation branch to provide a segmentation for affordance
(e.g., one-vs-all).  The \balg{Multi} segmentation branch grows in
direct proportion to the affordance rank quantity (here three). It is a
straightforward, brute force method used for comparison \citep{chu2019toward}. 
%
Another \textit{baseline} for segmentation employs a modified DeepLab
\citep{chen2018deeplab,chu2019toward}. 
DeepLab is a widely adopted segmentation network for semantic
segmentation, especially for urban street segmentation
\citep{ros2016synthia, Richter_2016_ECCV}. It is a representative
image-based segmentation approach 
for comparison.
%
With regards to \balg{\algName}, the implementation is also
modified to permit ranked affordances by using the KL-divergence instead
of the cross-entropy loss. However, due to the neural network's memory
footprint, only the regional attention mechanism is incorporated (no
attribute embedding).  We label it \balg{{\algNameKL}}.


%
%

\begin{figure}[t]
  \centering
  \includegraphics[height=0.65in]{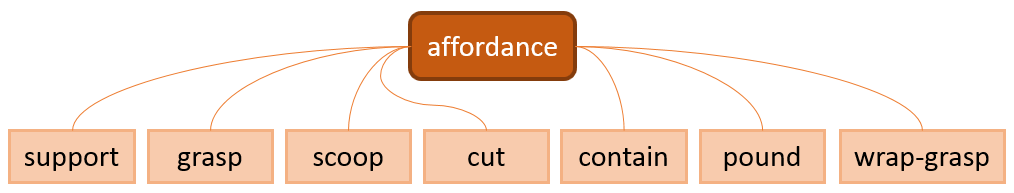}
  \caption{Affordances are used as attributes in an auxiliary module for
  per-candidate region multi-class classification in the UMD dataset.%
  \label{attribute_fig}}
\end{figure}
\begin{table}[t]
  \centering
  \caption{Images per Affordance \tablabel{attribute_table}}
  \small
   \addtolength{\tabcolsep}{-3pt}
  \begin{tabular}{ | l | l || l | l || l | l |}
   \hline
    affordance & images & affordance & images & affordance & images   
    \\ \hline \hline
    grasp      & 18235  & contain    & 7889   & wrap       & 5250  
    \\ \hline 
    cut        & 5542   & pound      & 2257   &            &       
    \\ \hline
    scoop      & 2869   & support    & 2317   &            &       
    \\ \hline  
  \end{tabular}
  \addtolength{\tabcolsep}{-3pt}
\end{table}

For the auxiliary task used during \balg{\algName} training,
we treat the original UMD affordance labels as attributes across
categories, see Fig. \ref{attribute_fig}. 
In total seven attributes (number of affordance) are defined for
representing the UMD tool dataset, 
as summarized in Table \tabref{attribute_table}. 


\subsection{Evaluation Metric}
To evaluate the affordance segmentation responses, derived from 
probability outputs over affordance classes, against ground truth
labels for each affordance, we adopt the weighted F-measures metric,
$F_{\beta}^{\omega}$, for the predicted masks:
\begin{equation} \label{F_measure}
  F_{\beta}^{\omega} = {(1 + \beta^{2})}{\frac{Pr^{\omega}\cdot Rc^{\omega}}{\beta^{2}\cdot Pr^{\omega}+Rc^{\omega}}}.
\end{equation}
where $Pr^{\omega}$ and $Rc^{\omega}$ are the weighed precision
and recall values, respectively \citep{margolin2014evaluate}.
Higher weights are assigned to pixels closer to foreground ground truth.
The weighted F-measures outputs lie in the range $[0,1]$.  
A second metric evaluates the prediction performance of
the rankings for multiple affordance on object parts and applies to
the KL-divergence trained network. 
It is the \textit{ranked} weighted F-measures metric,
$F_{\beta}^{\omega}$, 
\begin{equation} \label{ranked_F_measure}
  R_{\beta}^{\omega} = \sum_{r} \omega_{r}  F_{\beta}^{\omega}(r), 
  \quad \text{with} \quad \sum_r\omega_{r} = 1,
\end{equation}
where $\omega_{r}$ are the ranked weights contributing to the
weighted sum over the corresponding affordances
\citep{margolin2014evaluate}.
The top affordance receives the most weight and so on, per 
$\omega_{r} = 2^{-r} / \sum_{r^{\prime}}  2^{-r^{\prime}}$.
The ranked weighted F-measures outputs lie in the range $[0,1]$.

Evaluation results are obtained by running the same evaluation code
provided by UMD \citep{myers2015affordance}. 
All the parameters are the same.
There are two parameters associated with $\omega$ including $\sigma$ and $\alpha$.
Specifically, $\beta = 1$, $\sigma = 5$, and $\alpha = \frac{\ln{0.5}}{5}$.

\begin{figure*}[t]
  \centering
  \begin{tikzpicture}[inner sep=0pt,outer sep=0pt]
    \node (SF) at (0in,0in)
      {\includegraphics[height=1.65in,clip=true,trim=0in 0in 0in 0in]{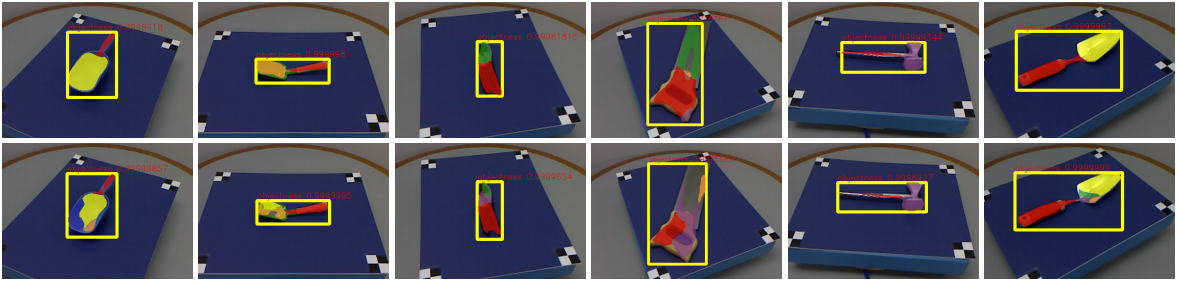}};
    
  \end{tikzpicture}
    \caption{Affordance segmentation results on UMD benchmark, where 
    color overlays represent affordance labels, 
	red: grasp; yellow: scoop; green:cut; dark blue: contain; 
	blue: wrap-grasp; orange: support; purple: pound. 
	{\bf Top:} results using \balg{\algName}; 
	{\bf Bottom:} results using \balg{Obj-wise} baseline.%
    \label{fig_UMD_results}}
\end{figure*}

\begin {table*}[t]
  \centering
  \newcommand{\TSfm}[1][2.0ex]{\rule{0pt}{#1}}
  \caption {Affordance Segmentation Performance On UMD Dataset (novel
    category).\tablabel{fmeasure} }
  \small
  \addtolength{\tabcolsep}{3pt}
  \begin{tabular}{ | l | c | c |  c | c | c | c | c || c |}
    \hline
    & \multicolumn{8}{c|}{ \bf{weighted F-measures}} \\ \hline
    & grasp    & cut & scoop       & contain & pound   &  support & w-grasp 
    & average  \\ \hline
    \cite{Rezapour_Lakani2018}   
      & 0.46  & 0.30 &  0.22   & 0.47   & 0.08    &  0.03   &  0.47  & 0.29 
      \TSfm \\ \hline \hline
    DeepLab \citep{chu2019toward}	   
      & 0.55  & 0.30 &  0.36   & 0.58 & 0.42    &  0.22   &  0.93   & 0.48 
      \TSfm \\ \hline 
    Obj-wise  \citep{chu2019toward}
      & 0.61  & 0.37 &  0.60   & 0.61   & 0.81    &  0.59   &  0.94  & 0.64 
      \TSfm \\ \hline 
    KLdiv-1  \citep{chu2019toward}
      & 0.54  & 0.31 &  0.39   & 0.63   & 0.55    &  0.75   &  0.92  & 0.58 
      \TSfm \\ \hline 
    Multi-1 \citep{chu2019toward}
      & 0.56  & 0.35 &  0.53   & 0.60   & 0.69    &  0.68   &  0.92  & 0.62 
      \TSfm \\ \hline \hline
      {\algName}                                
      & 0.60  & 0.37 &  0.60   & 0.61   & 0.80    &  0.88   &  0.94  &\bf{0.69} 
      \TSfm \\ \hline
      {\algNameKL}-1
      & 0.54  & 0.37 &  0.42   & 0.62   & 0.63    &  0.87   &  0.92  &\bf{0.63} 
      \TSfm \\ \hline
  \end{tabular}
  \addtolength{\tabcolsep}{-3pt}

  \centering
  \vspace*{0.1in}
  \newcommand{\TSrf}[1][2.0ex]{\rule{0pt}{#1}}
  \caption {Affordance Ranking Performance On UMD Dataset (novel category) \tablabel{ranked_fmeasure} }
  \small
  \addtolength{\tabcolsep}{3pt}
  \begin{tabular}{ | l | c | c |  c | c | c | c | c || c |}
    \hline
    & \multicolumn{8}{c|}{ \bf{ranked weighted F-measures}} \\ \hline
                & grasp   & cut  & scoop & contain  & pound    & support  & w-grasp & average \\ \hline 
    HMP \citep{myers2015affordance}       
      & 0.16   & 0.02 & 0.15 & 0.18  &  0.02   & 0.05  & 0.10   &  0.10  
      \TSrf \\ \hline
    SRF \citep{myers2015affordance}       
      & 0.05   & 0.01 & 0.04 & 0.07  &  0.02   & 0.01  & 0.07   &  0.04 
      \TSrf \\ \hline
    VGG \citep{sawatzky2017weakly}        
      & 0.18   & 0.05 & 0.18 & 0.20  &  0.03   & 0.07  & 0.11   &  0.12  
      \TSrf \\ \hline
    ResNet \citep{sawatzky2017weakly}	    
      & 0.16   & 0.05 & 0.18 & 0.19  &  0.02   & 0.06  & 0.11   &  0.11  
      \TSrf \\ \hline
    \cite{Rezapour_Lakani2018}    
      & 0.19   & 0.18 & 0.28 & 0.32  &  0.08   & 0.11  & 0.32   &  0.21  
      \TSrf \\ \hline \hline
    DeepLab \citep{chu2019toward}       
      & 0.299   & 0.172 & 0.101 & 0.223 &  0.056   & 0.037  & 0.531   &  0.203   
      \TSrf \\ \hline 
    KLdiv  \citep{chu2019toward}
      & 0.322   & 0.175 & 0.178 & 0.232  &  0.093   & 0.096  & 0.525   &  0.232
      \TSrf \\ \hline
    Multi \citep{chu2019toward}
      & 0.336   & 0.200 & 0.211 & 0.247  &  0.082   & 0.109   & 0.526   &\bf{0.244}
      \TSrf \\ \hline \hline
    {\algNameKL}                                 
      & 0.331   & 0.184 & 0.183 & 0.254  &  0.101   & 0.103   & 0.533   &\bf{0.241}
      \TSrf \\ \hline
  \end{tabular}
  \addtolength{\tabcolsep}{-3pt}
  
\end {table*}

\subsection{Benchmarking on UMD Novel Objects}
The traditional benchmarking scheme for the UMD dataset is to perform
an image-split test, where all object categories are represented. 
Affordance evaluation is performed only for the known object categories.
Top performance for the image-split test lies in the range of 0.733 to
0.799 for the weighted F-measures
\citep{nguyen2016detecting,do2018affordancenet,chen2018deeplab}.
The category-split test, where some object categories are excluded is a
more difficult problem since top performing methods rely on the
object category prediction to provide a prior on the potential affordances. 
The weighted F-measures decrease for the category-split test and there
are less reported evaluations. Especially when considering evaluation
for weighted F-measures and ranked weighted F-measures. 
For example, DeepLab performance drops by {34.5\%} (to 0.48 from 0.733). 
Since this study aims to explore affordance recognition in the absence
of object category knowledge, the category-split test case is performed
for single affordance and ranked affordance prediction.  Published
baselines are included in the benchmarking results when available. 
When presenting the results, the table contents will be organized
according to (i) published results, (ii) strong baselines created in
earlier efforts \citep{chu2019toward}, and (iii) the current results for
\balg{\algName}.


\subsubsection{Novel Category Affordance Prediction.}
Qualitative results of \balg{\algName} are depicted in the top row
Fig.~\ref{fig_UMD_results}, which has color overlays of the 
affordance predictions on the objects.  The bottom row shows the same
for the best performing baseline approach per Table \tabref{fmeasure}, 
which is the \balg{Obj-wise} implementation. 
\balg{\algName} improves the consistency of affordance segmentation as
seen by less oversegmentation and more spatially uniform affordance labels.

%
Quantitative evaluation using the weighted F-measure for the UMD
benchmark is found in Table \tabref{fmeasure}, with available published
outcomes for the category-split test. For the multi-affordance
\balg{KLdiv} baseline, the top ranked affordance is taken as the
affordance output. Hence the label \balg{KLdiv-1}.  
Likewise the single affordance version of \balg{\algNameKL} outputs only the
top ranked affordance label (denoted \balg{\algNameKL-1}).
\balg{\algName} has the strongest performance, with the \balg{Obj-wise}
baseline next. 
Compared to the best published result, the proposed approach achieves a 43\%
improvement (0.48 to 0.69). Compared to the strong baseline, it
achieves 7\% improvement over \balg{Obj-wise}.  
Meanwhile \balg{\algNameKL-1} has a 1.5\% drop in performance relative
to \balg{Obj-wise}, while the non-attention version \balg{KLdiv-1} has a
9.4\% drop in performance. 
The attribute and attention modules lead to improved performance in the
primary affordance segmentation outcomes.

\begin {table*}[t]
  \centering
  \caption {Ablation Study \tablabel{ablation_fmeasure} }
  \small
  \setlength{\tabcolsep}{8.4pt}
  \begin{tabular}{ | l || c | c || c | c |  c | c | c | c | c || c |}
    \hline
    & \multicolumn{2}{c||}{module} & \multicolumn{8}{c|}{ \bf{weighted F-measures}} \\ \hline
             &  attent & attri  & grasp   & cut  & scoop & contain  & pound    & support  & w-grasp & average \\ \hline

    Obj-wise  &          &              & 0.61  & 0.37 &  0.60   & 0.61   & 0.81    &  0.59   &  0.94   & 0.64  \\ \hline
    Obj-wise  &          &  \ding{51}   & 0.62  & 0.33 &  0.51   & 0.61   & 0.81    &  0.79   &  0.94   & 0.66  \\ \hline     
    Obj-wise  &\ding{51} &              & 0.60  & 0.40 &  0.67   & 0.60   & 0.78    &  0.75   &  0.94   & 0.68 \\ \hline
    Obj-wise  &\ding{51} &  \ding{51}   & 0.60  & 0.37 &  0.60   & 0.61   & 0.80    &  0.88   &  0.94   &\bf{0.69}\\

    \hline
  \end{tabular}
  \vspace*{-0.05in} 
\end {table*}

\subsubsection{Novel Category Affordance Ranking.}
Affordance ranking with ranked weighted F-measures is reported in Table
\tabref{ranked_fmeasure}\,. 
This test is harder, due to the metric heavily penalizing incorrect
rankings, which compresses the score output values.
The results show that \balg{\algName} outperforms the existing published
approaches and most strong baselines for novel object categories.
Compared to the most recent published results \citep{Rezapour_Lakani2018}
and to the strong baseline DeepLab, \balg{\algName} improves by 14\% and
20\%, respectively.  Evaluation of \balg{\algNameKL} relative to
\balg{KLdiv} and \balg{Multi} shows that \balg{\algNameKL} outperforms
\balg{KLdiv} and almost matches \balg{Multi}. 
%
Recall that the implementation of \balg{Multi} \citep{chu2019toward} in
Table \tabref{ranked_fmeasure} is especially designed for affordance
ranking by adopting multiple afforance branches for each rank output.
Consequently it scales in parameter size linearly with the rank
quantity (three). The \balg{KLdiv} approach does not impact the
parameter size of the affordance ranking branch (compared to
cross-entropy loss). The \balg{\algNameKL} approach increases by less
than 1\% the branch network size.
%
%
%
%
%
%
The gap between \balg{KLdiv} ($0.2315$) and \balg{Multi} ($0.2444$)
is 5.3\%.  In contrast, for a less than 1\% increase in branch
parameters, \balg{\algName} ($0.2414$) reduces the performance drop to
1.2\%. In essence, for a small increase in size the attention module
almost matches the performance of the brute force (one-vs-all) network.
Given that \balg{\algNameKL-1} outperforms \balg{Multi} for the primary
affordance, per Table \tabref{fmeasure}, the main output differences lie
with the secondary and tertiary affordances.

\subsection{Ablation Study}
Table \tabref{ablation_fmeasure} shows the ablation study results for 
\balg{\algName}, where the baseline network is the \balg{Obj-wise}
network (first row).  The second and third rows quantify the
improvements gained from regional attention and attribute learning,
respectively. 
The last row reports the performance with both attention and attribute
learning together, which achieves the best result. 
Each design is independently trained with ImageNet pre-trained weights,
instead of finetuing one-by-one. 
While a 3.7\% (0.770 to 0.799) improvement in \cite{do2018affordancenet} was
regarded as reasonable achievement on UMD image-split benchmark, the ablation
study shows 3.1\% and 6.2\% improvements by introducing attribute learning
and ROI-based attention individually. 
Furthermore, the current gap between the best performing object-aware
approach (0.799 for \cite{do2018affordancenet}) and the best performing
object-agnostic approach (0.48 for DeepLab) has been
reduced from a 40.0\% drop to a 13.6\% drop and lies close to the
lower-end of the range for state-of-the-art object-aware methods
(0.733-0.799), e.g., within 6\%.
The next section, which performs manipulation tests, will further
explore how this difference manifests when considering task-relevant
manipulation activities, from simply grasping and picking up an object,
to performing affordance aware manipulation tasks.

\subsection{Affordance Detection across Datasets}
\begin{figure}[t]
  \centering
  \begin{tikzpicture}[inner sep=0pt,outer sep=0pt]
    
  \node (SF) at (0in,0in)
    {\includegraphics[height=1.45in,clip=true,trim=0in 0in 0in 0in]{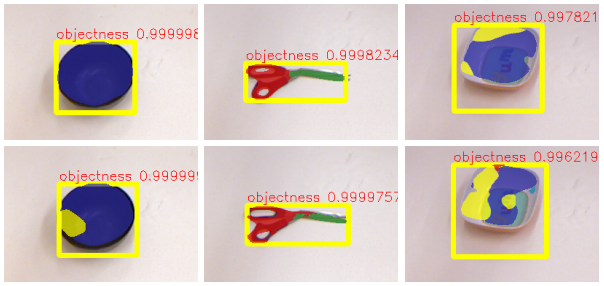}};
  \end{tikzpicture}
  \caption{Comparison on Cornell dataset.
    {\bf Top:} detection results of \balg{\algName}  method. 
    {\bf Bottom:} \balg{Obj-wise} model.%
    \label{fig_cornell_results}}
\end{figure}

The last vision-only test demonstrates generalizability across datasets. 
\balg{\algName} trained on the novel-category split of the UMD dataset is
applied to the Cornell dataset \citep{cornell2013}. 
The Cornell dataset consists of 885 images of 244 different objects for
learning robotic grasping. Each image is labelled with multiple ground
truth grasps.
Though affordance masks are not available for quantitative evaluation
with $F_{\beta}^{\omega}$ and ranked $F_{\beta}^{\omega}$ metrics,
visualizations of qualitative results and comparisons are presented in
Fig. \ref{fig_cornell_results}.
Similar to Fig.~\ref{fig_UMD_results}, the outcomes here have more
consistent and continuous affordance segmentations for \balg{\algName}
versus \balg{Obj-wise}.

\section{Manipulation Experiments and Methodology}
\begin{figure}
  \centering
  \begin{tikzpicture}[inner sep=0pt,outer sep=0pt]
    \node (SF) at (0in,0in)
      {\includegraphics[height=2.4in,clip=true,trim=0in 15in 0in 9in]{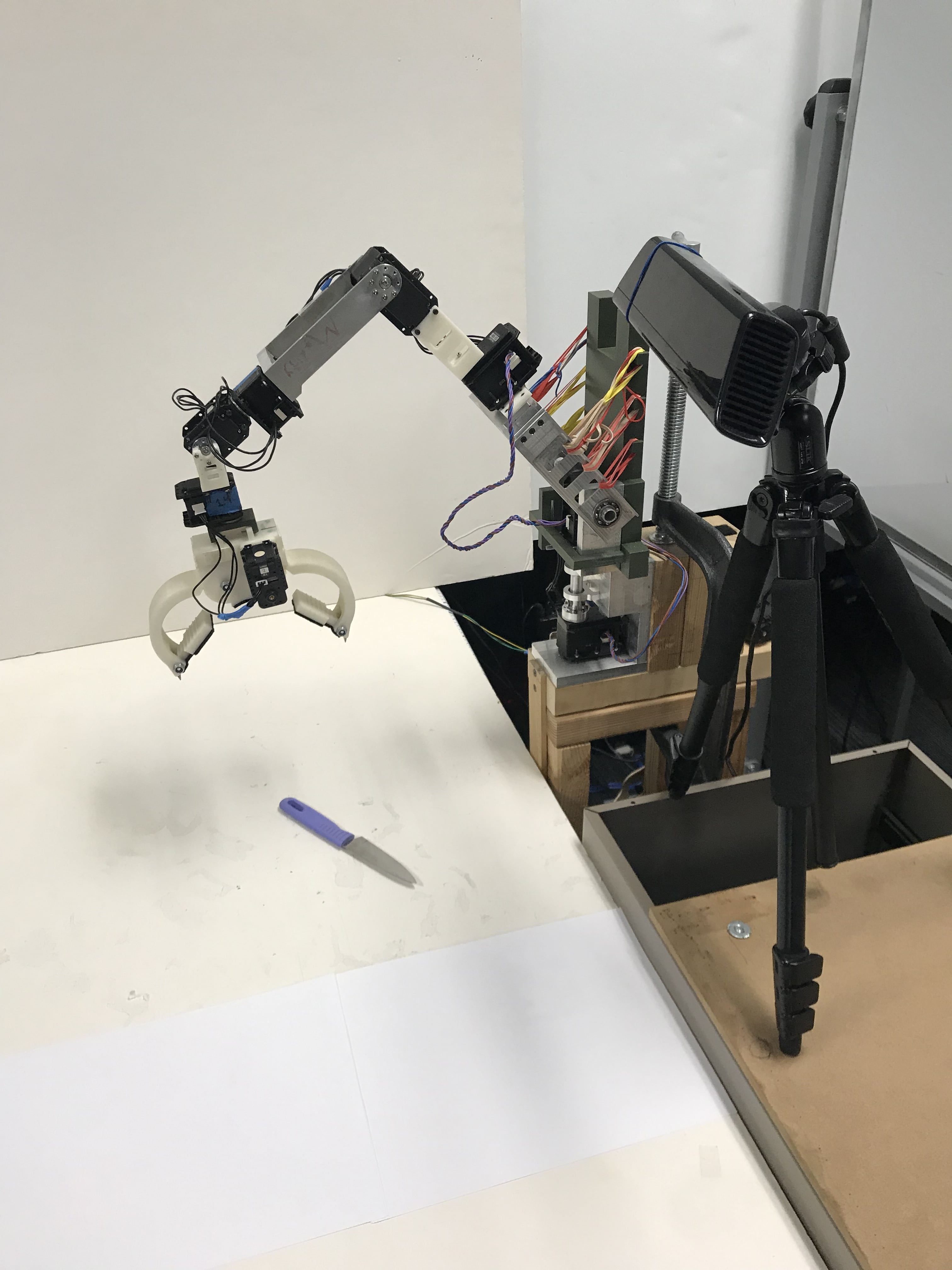}};
  \end{tikzpicture}
  \caption{Experimental setup for eye-to-hand physical manipulation,
  with a 7 DoF manipulator and a Microsoft Kinect RGB-D sensor.%
  \label{exp_setting}}
\end{figure}

Beyond vision-only benchmarking, embodied robotic manipulation with
affordance detection is tested and evaluated.  The physical manipulation
tests use a custom-built 7-DoF robotic manipulator and a Microsoft
Kinect using an eye-to-hand configuration (see Fig. \ref{exp_setting}).
In the experiment setting, the sensor input includes depth information.
Image areas with target affordances are mapped to 3D space for
manipulation planning and execution. 
Multiple scenarios and possible applications are validated. This
section describes the experimental methodology of the five test scenarios, 
with the subsequent section covering the results.
%

\subsection{Affordance on Seen Categories} \label{exp_Seen_Categories}
\begin{figure}[t]
  \centering
  \begin{tikzpicture}[inner sep=0pt,outer sep=0pt]
    \node (SF) at (0in,0in)
      {\includegraphics[width=1.1in,clip=true,trim=0in 7.5in 0in 1.5in]{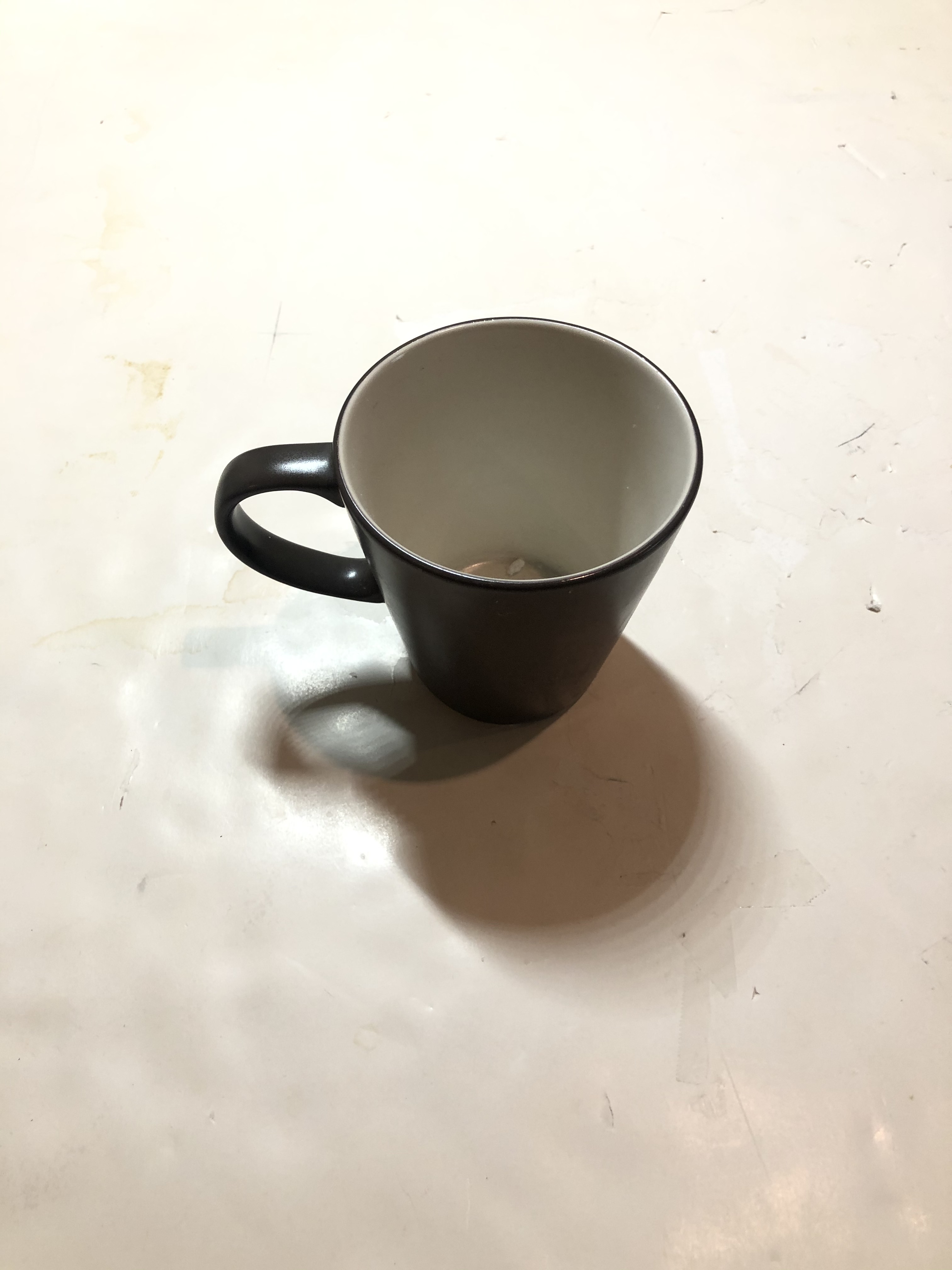}};
    \node[anchor=south west, xshift = 0.05cm] (MF) at (SF.south east)
      {\includegraphics[width=1.1in,clip=true,trim=0in 7.5in 0in 1.5in]{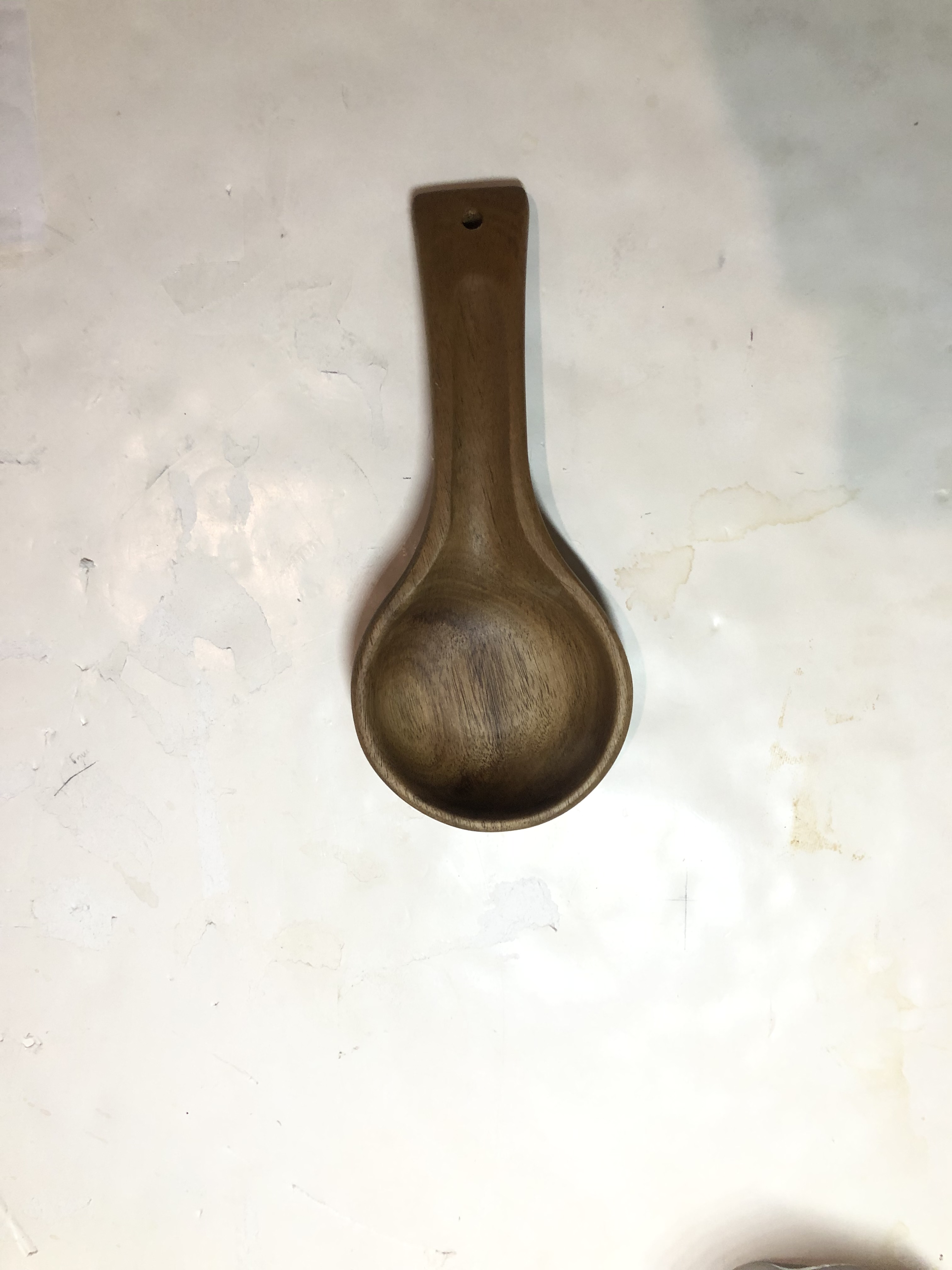}};
    \node[anchor=south west, xshift = 0.05cm] (QF) at (MF.south east)
      {\includegraphics[width=1.1in,clip=true,trim=0in 2.5in 0in 6.5in]{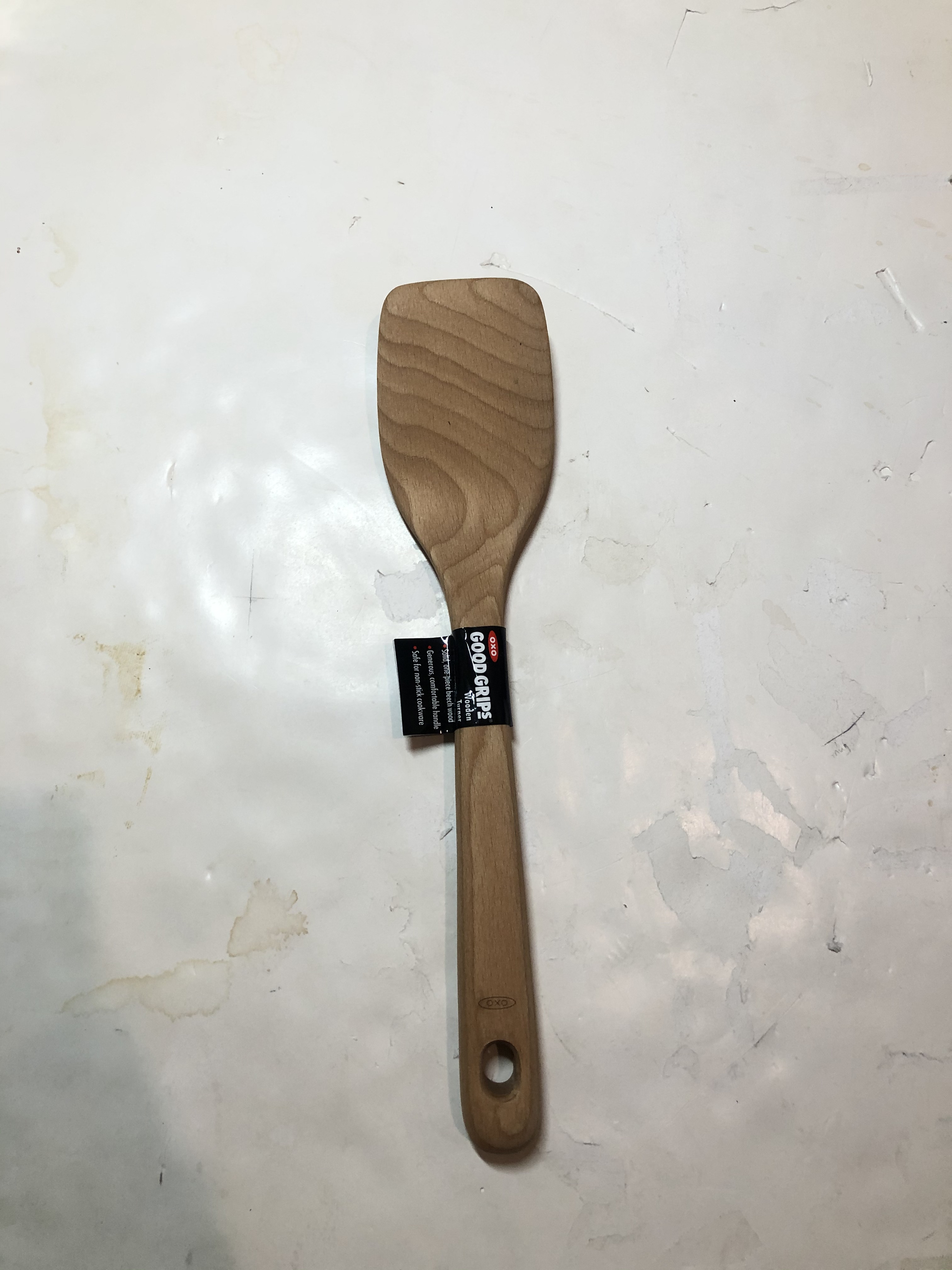}};

    \node[anchor=north west, xshift = 0cm, yshift = -0.05cm] (NF) at (0in,0in) at (SF.south west)
      {\includegraphics[width=1.1in,clip=true,trim=0in 2in 0in 1in]{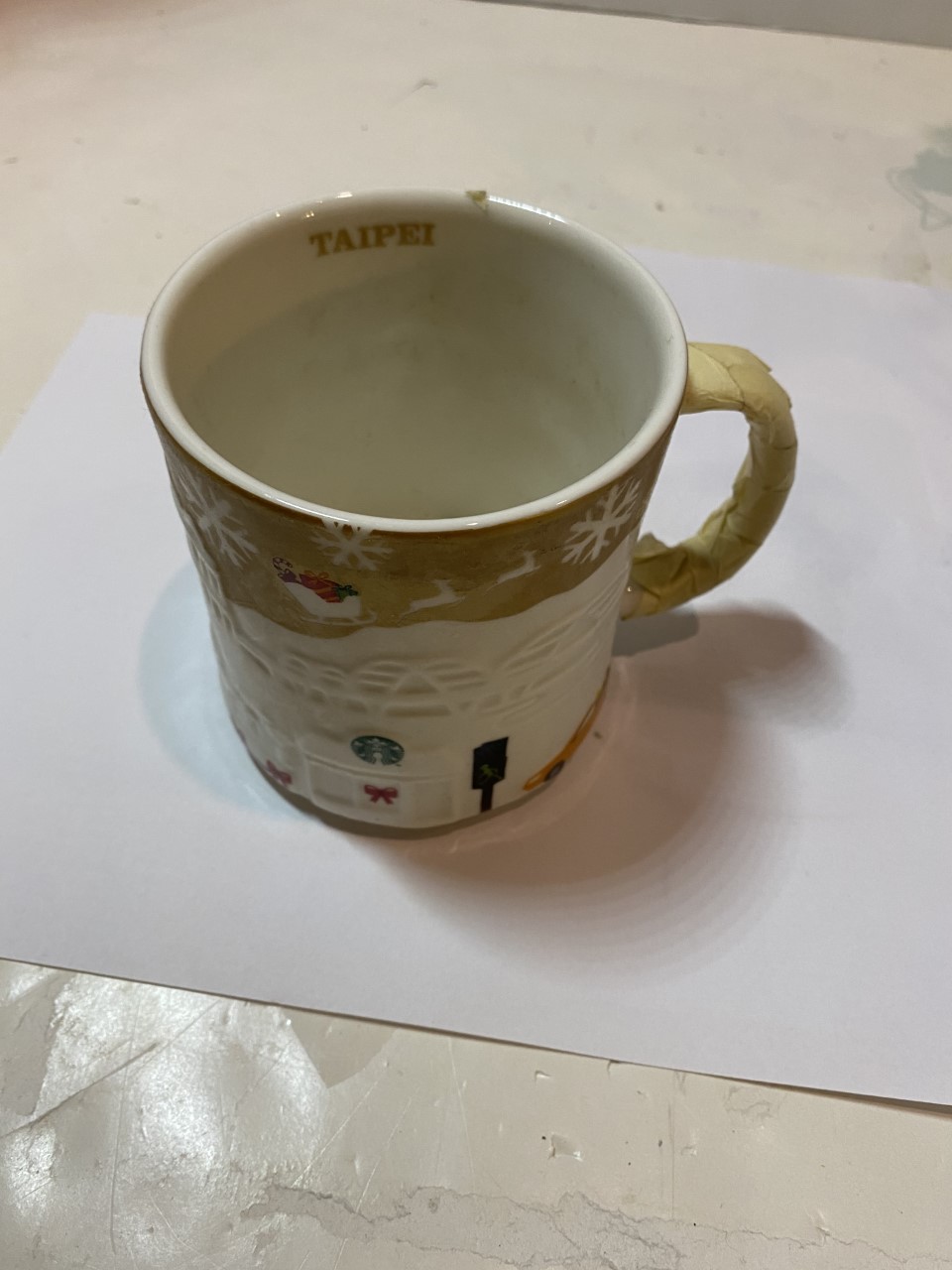}};
    \node[anchor=south west, xshift = 0.05cm] (OF) at (NF.south east)
      {\includegraphics[width=1.1in,clip=true,trim=0in 6.5in 0in 3in]{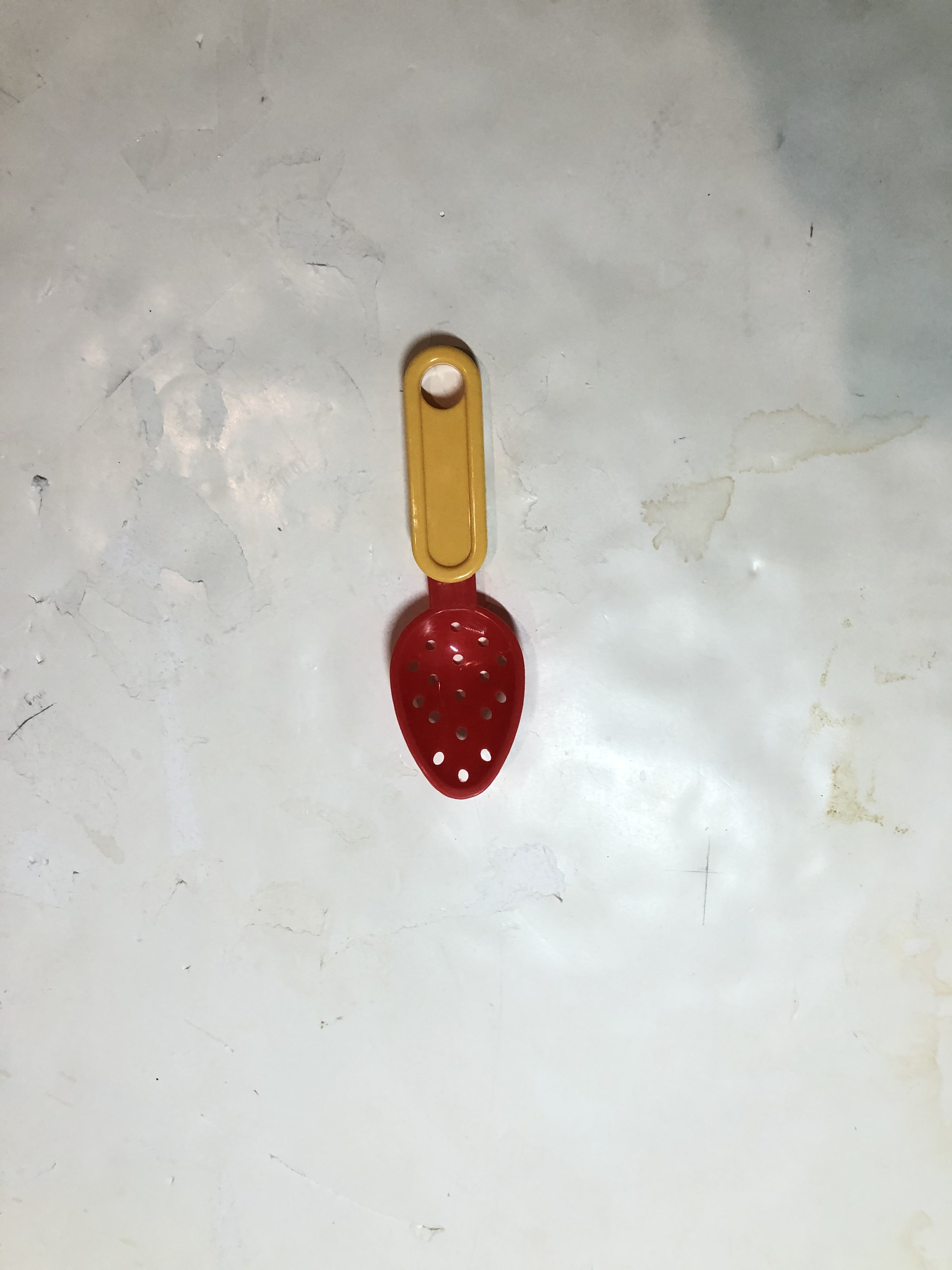}};
    \node[anchor=south west, xshift = 0.05cm] (PF) at (OF.south east)
      {\includegraphics[width=1.1in,clip=true,trim=0in 6.5in 0in 3in]{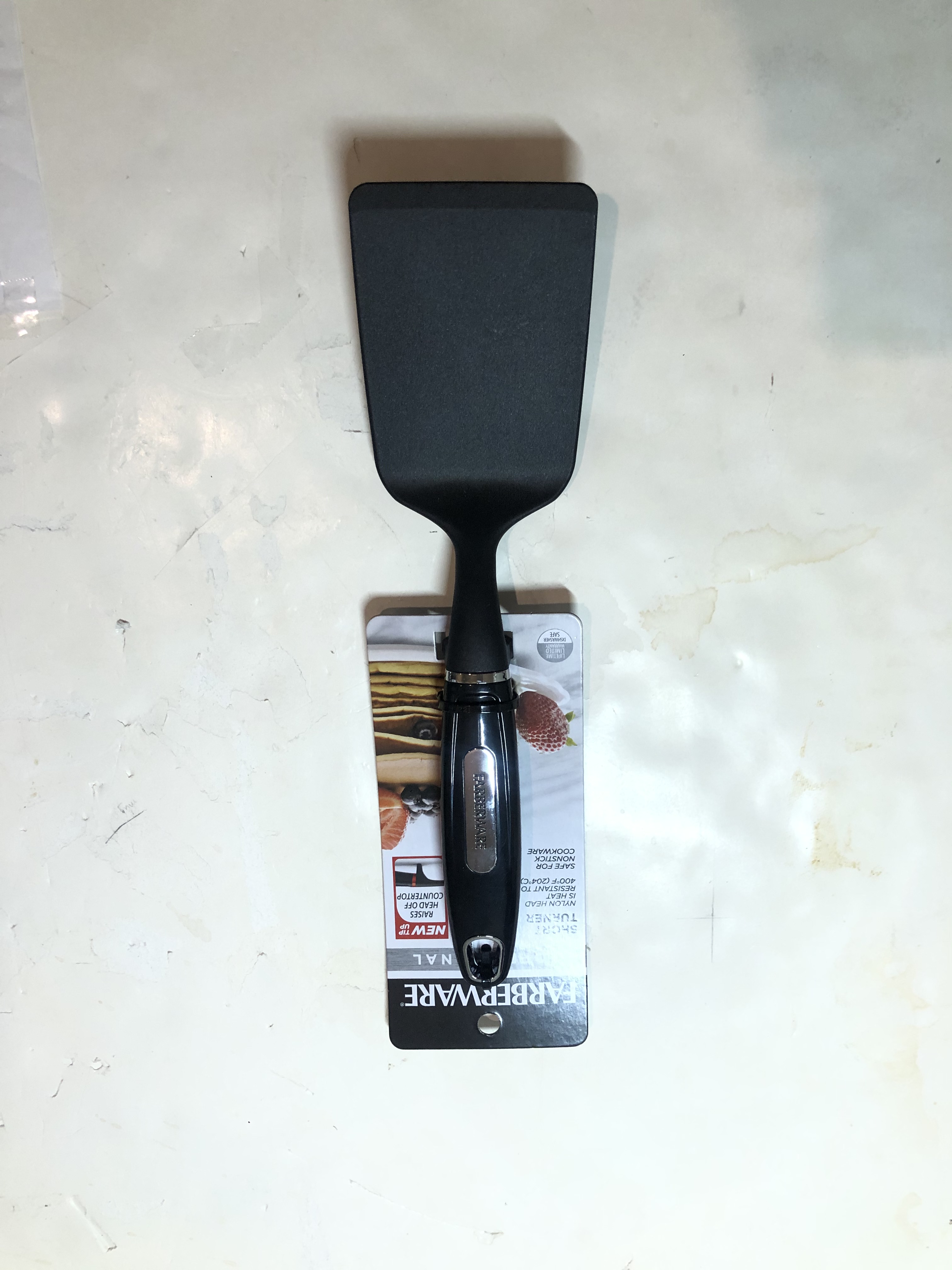}};

    \node[anchor=north west,xshift=3pt,yshift=-3pt] at (SF.north west) {(a)};
    \node[anchor=north west,xshift=3pt,yshift=-3pt] at (MF.north west) {(b)};
    \node[anchor=north west,xshift=3pt,yshift=-3pt] at (QF.north west) {(c)};
    \node[anchor=north west,xshift=3pt,yshift=-3pt] at (NF.north west) {(d)};
    \node[anchor=north west,xshift=3pt,yshift=-3pt] at (OF.north west) {(e)};
    \node[anchor=north west,xshift=3pt,yshift=-3pt] at (PF.north west) {(f)};
   
    \end{tikzpicture}
    \captionof{figure}{Examples of similar (top row) and dissimilar 
    objects (bottom row) relative to the UMD dataset.  
    (a) and (d): Most mugs are small and have no or few visual patterns;
        the dissimilar mug is large and has patterns.
    (b) and (e): Most spoons large serving spoons; the dissimilar spoon
        is a small toy. 
    (c) and (f): Most turners are made of wood or steel; 
        the dissimilar turner is made of plastic.%
    \label{fig_simple_movement}}
\end{figure}
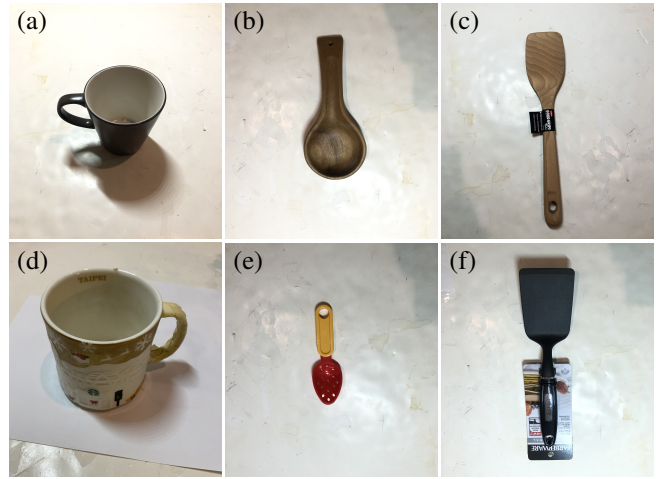

%
%
%
As in \cite{do2018affordancenet}, commonly seen \textit{grasp} and
\textit{contain} affordances are first examined. In addition, the
\textit{support} affordance is included.
All the objects in this experiment are selected from categories in UMD dataset,
and thus can be recognized by \textit{AffNet} (trained on UMD dataset). Two instances are picked in each category, where one is similar to instances in UMD dataset, and the other is dissimilar. Examples of selected similar and dissimilar objects are shown in Fig. \ref{fig_simple_movement}.
The experiment is mainly designed for benchmarking the proposed
category-agnostic detector against the state-of-the-art affordance
detection \citep{do2018affordancenet}, which is trained with object prior but limited to the training categories during deployment. 

For experiments involving the \textit{grasp} affordance, the grasp
center is computed by averaging the \textit{grasp}-able pixels, with grasping orientation determined by fitting a line to predicted pixels. 
Since the camera is set near top-down view, the orientation is then corrected by the relative orientation between the input sensor and working space.  
%
%
For the \textit{contain} and \textit{support} affordances, a small cuboid is placed into (or onto) an object
predicted as \textit{contain}-able (\textit{support}-able) with the robotic arm. As with
the \textit{grasp} affordance, the location is determined by averaging
the pixels predicted as \textit{contain}-able (\textit{support}-able).
The evaluation metric is discussed in the next section.
%
%

%

\subsection{Affordance with Multiple Objects} \label{exp_Multiple}
More common scenarios usually involve multiple objects in a scene. 
The workspace may contain task-irrelevant objects. For instance, the
robotic arm is required to grasp a knife while a plate and cup are in
the scene. The manipulator is asked to put an object into the cup while
knife and spoon are around. 
%
%
Though the proposed method is trained on UMD dataset with single object
annotated in each image, the trained model is readily available to
detect multiple objects with corresponding affordance maps.
This experiment replicates the first one,
but uses a dissimilar set of objects, such that there are no common
affordances.
%
For each trial, multiple objects (at least 3) including the target
object (with the target affordance) are presented and randomly placed in a
visible and reachable area.  The experiment tests affordance recognition
correctness and consistency in the presence of nuisance categories. If
another object is misattributed the target affordance, then the planner
may output an incorrect action sequence.  

\subsection{Affordance on Unseen Categories} \label{exp_Unseen_Categories}
This experiment tests generalizability to unseen categories. Having this
capability can mitigate labor-intensive pixel-wise and bounding box
annotations. 
While \textit{AffNet} achieves slightly better vision
performance, it is limited to the categories existing in training set
(17 object categories in UMD dataset). 
Therefore, when replicating the first experiment,
it modifies the object set to draw from objects not seen in the UMD
object categories. 
%
%
%
%

\subsection{Affordance for Task-Oriented Grasping \subseclabel{TOGrasp}}
Affordances such as \textit{grasp}, \textit{contain} and
\textit{support} are usually considered independently from the
manipulation perspective. In contrast, some affordances like
\textit{pound} and \textit{cut} are commonly utilized in combination
with \textit{grasp} and a corresponding action primitive. 
Task-oriented grasping experiments were designed to evaluate these affordances.
Task-oriented grasping requires identifying a proper graspable part and
functional part in order to accomplish the task. In this section, two
task-oriented grasping experiments including \textit{pound peg into slot} task
and \textit{cut through string} task are described.  

\begin{figure}[t]
  \centering
  \begin{tikzpicture}[inner sep=0pt,outer sep=0pt]
    
    \node (SF) at (0in,0in)
      {\includegraphics[height=1.70in,clip=true,trim=0in 0.5in 0in 1.5in]{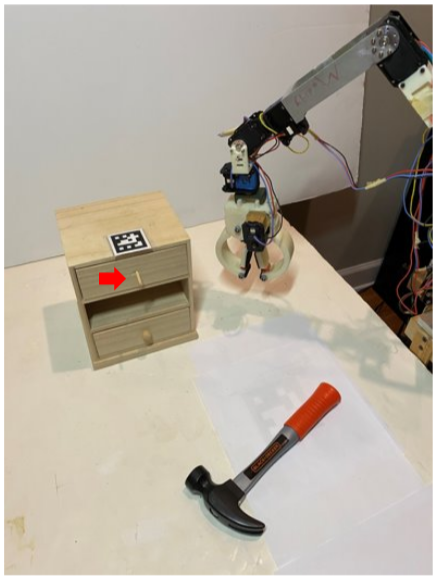}};
    \node[anchor=south west, xshift = 0.05cm] (MF) at (SF.south east)
      {\includegraphics[height=1.70in,clip=true,trim=0in 0.5in 0in 1.5in]{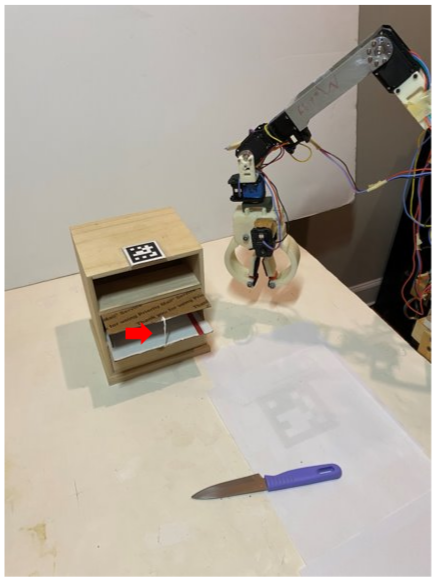}};
    \node[anchor=south west, xshift = 1pt] (NF) at (0in,0in) at (SF.south west)
      {\includegraphics[height=.6in,clip=true,trim=1.0in 1in 0.2in 2.0in]{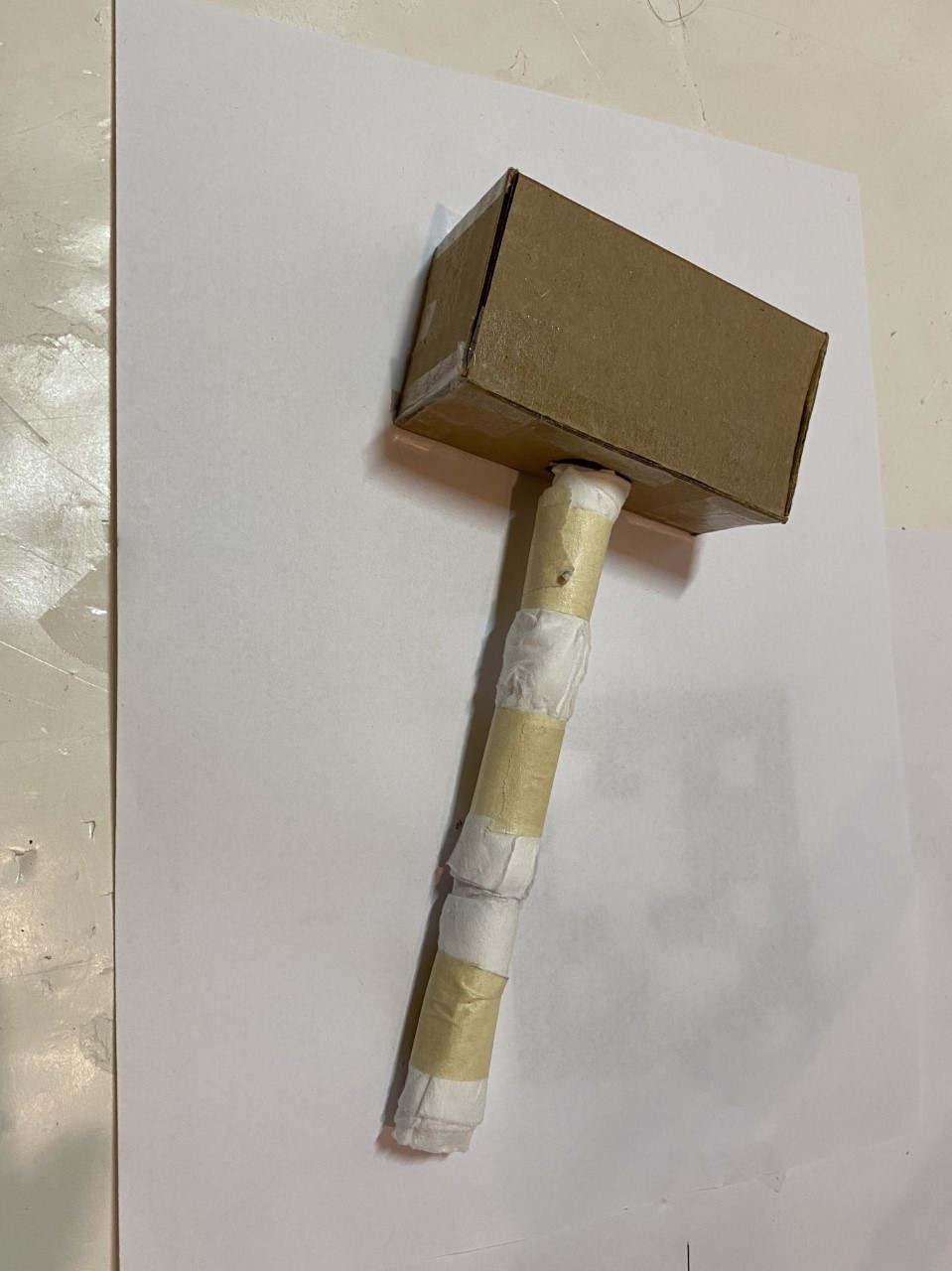}};
    \node[anchor=south west, xshift = 1.5pt] (OF) at (MF.south west)
      {\includegraphics[height=0.6in,clip=true,trim=2.5in 3.0in 1in 3.0in]{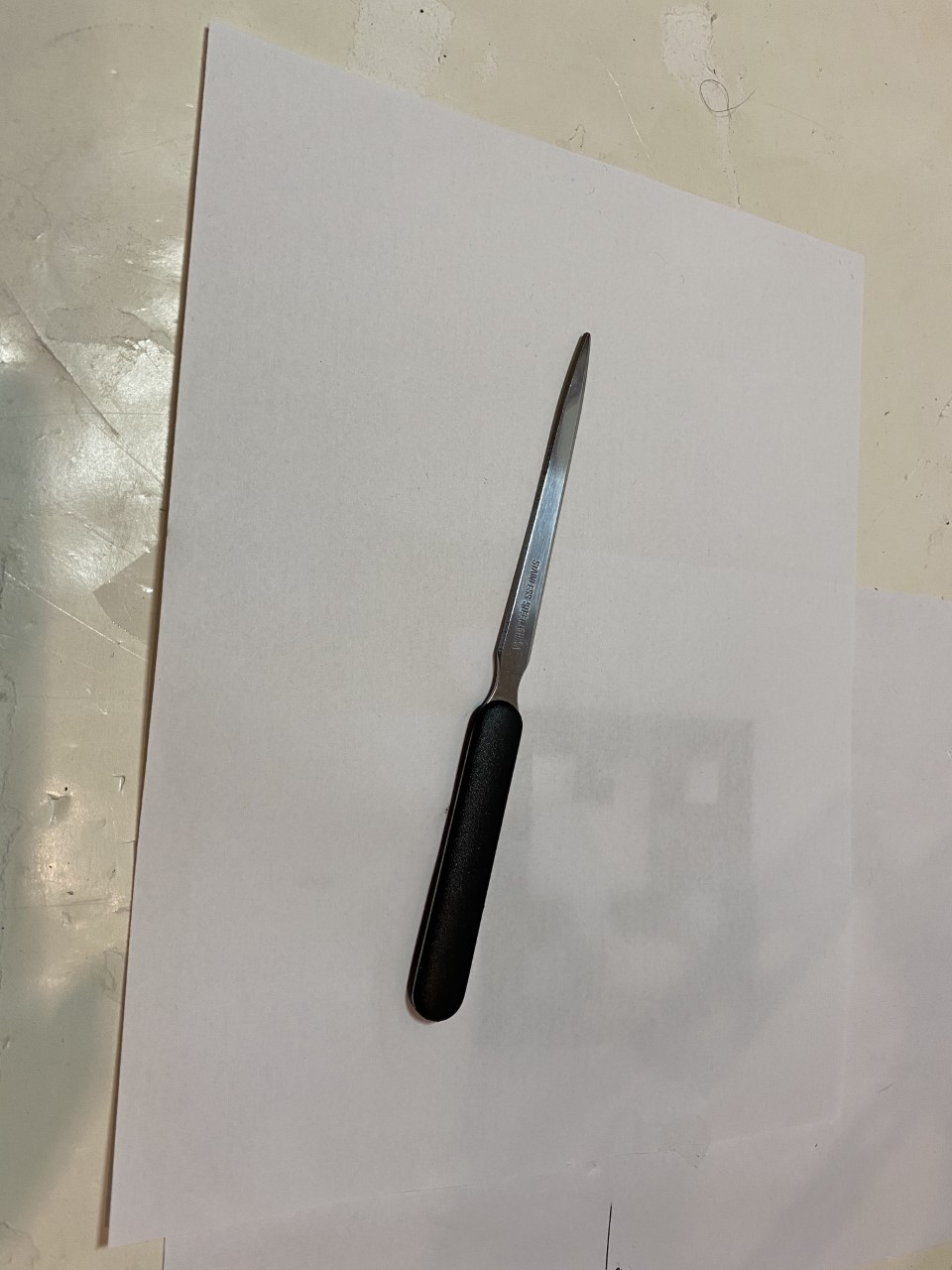}};

    \node[anchor=north west,xshift=1pt,yshift=-2pt] at (SF.north west) {(a)};
    \node[anchor=north west,xshift=1pt,yshift=-2pt] at (MF.north west) {(b)};
    \node[anchor=north west,xshift=1pt,yshift=-2pt] at (NF.north west) {(c)};
    \node[anchor=north west,xshift=1pt,yshift=-2pt] at (OF.north west) {(d)};
   
  \end{tikzpicture}
  \captionof{figure}{Illustration of task-oriented manipulation settings.
    \textbf{(a)} \textit{peg into slot task}: A peg (see red arrow) is
      half-way through the slot. The manipulator needs to grasp the tool
      (hammer or tenderizer) and pound the peg fully into the slot.
    \textbf{(b)} \textit{cut through string task}: A string made by
      tissue is attached vertically (see red arrow). The manipulator
      needs to grasp the tool (knife or letter opener) and fully cut off
      the string.  
    \textbf{(c)} Custom-made toy tenderizer. 
    \textbf{(d)} Letter opener. 
            }
            \label{fig_task_oriented}
\vspace*{0.25em}
      \centering
      {\includegraphics[width=0.49\columnwidth,clip=true,trim=0in 3.15in 0in 0in]{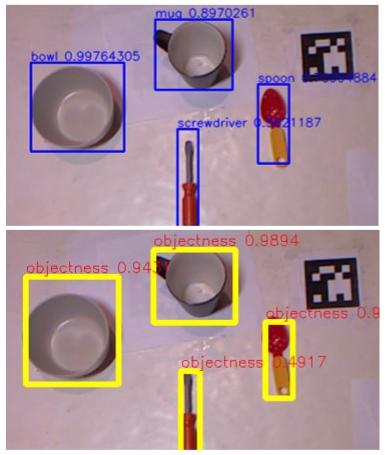}}
      {\includegraphics[width=0.49\columnwidth,clip=true,trim=0in 0in 0in 3.15in]{boundingbox.png}}
   
  \caption{
    {\bf Left:} Candidate bounding boxes (in blue) predicted by a
        pre-trained object detector. 
    {\bf Right:} Candidate bounding boxes (in yellow) predicted by
        proposed affordance detector. 
    The detectors are indepdently trained yet predict candidates that
    are close in locations and similar in size.  
    \label{fig_bnbbox}}
\end{figure}

The \textit{pound peg into slot} task is setup as shown in
Fig.~\ref{fig_task_oriented}(a). The peg is initially inserted half-way
into a slot and placed within reach of the manipulator. The manipulator is
required to detect and grasp the tool, then use it to pound the peg
fully into the slot (three strikes are programmed for each trial). 
Solving it involes using the \textit{grasp} affordance to grab the object.
The distance between the grasping point and the functional point (where
to pound on the hammer head) is computed from the affordance map and
depth image. The Aruco marker at the top of the box is for detecting
the relative location of the peg. 

The second task, \textit{cut through string}, is shown in Fig.
\ref{fig_task_oriented}(b). The string is a twisted tissue fixed
vertically between two horizontal surfaces and taped to them at each
end.  The manipulator is required to find and grasp the handle of the
tool, and cut the string off horizontally using a programmed wrist
motion (three slice motions are programmed for each trial). 
Again, the Aruco marker is for detecting the location of the string,
while the grasping point and cutting point is computed through the
affordance map and depth image.
%

\subsection{Affordance with modified PDDL pipeline} \subseclabel{mPDDL}
Category-agnostic affordance detection generalizes learned affordances to unseen categories by detecting objectness instead of identifying specific object classes. However, knowing object classes may be required in some scenarios such as grasping a tool from a toolbox, or pouring water into a cup next to a bowl. 
This can be completed by using the object-agnostic affordance detector
with a pre-trained (affordance-agnostic) object detector. In
this experiment, an object detector (Faster-RCNN with a VGG-16 backbone)
is trained on 21 objects (items commonly seen at home or on office table). 
As shown in Fig. \ref{fig_bnbbox}, though the affordance and
object detectors are trained independently, they both predict similar
candidate bounding boxes in our setting. The predictions from the two
detectors are associated through overlapping bounding box locations
and similar sizes.
To increase the complexity of the task, this experiment looks at
compound task specifications where the objects have coupled
or interacting specifications.  It also involves scene states that have
objects on or in other objects, which result in reduced detection and
recognition probability. Missed detections or recognitions mean the
state cannot be established from the recovered visual information.
Keeping track of the symbolic state and its spatial parameters via the
\textit{state keeper} is necessary. 
These tasks require the \textit{state keeper} module to transfer the terminal state of the prior
action to the initial state of the subsequent action. An example would
be the state \textit{plate contains spoon} as well as the location of 
the spoon in scene.
Four experiments are designed to evaluate the full pipeline, which
includes the \textit{pre-trained object detector} and the \textit{state
keeper}.

The first two experiments, \textit{pick knife/spoon into bowl} and
\textit{select trowel/spoon to scoop beans} examine the combined use
of object detection and affordance recognition to select from objects
in the view to accomplish the tasks.  Additionally, it is designed to
have a nuisance object in the scene with similar affordances.
Consider the first experiment.  Though both objects are in the scene, at
the commencement of each trial the \textit{knife} or \textit{spoon} is
randomly selected to be the object to place into the bowl so that the
other becomes the nuisance object. 
The object detector is required to identify the chosen object, while the
affordance recognition process is needed to \textit{grasp} it.
Likewise, the pipeline should correctly interpret the bowl in the scene.
It implicitly tests the ability of the two detectors to identify the
same object regions, as in Fig. \ref{fig_bnbbox}, and to recognize their
object types and affordances. The equivalent setup and objective applies
to the second experiment, \textit{select trowel/spoon to scoop beans}. 

For some deployments, the robot manipulator may not get a single task to
perform, but may get a series of tasks spaced out in time, that may or
may not be related. Awareness of prior activities is important when the
resulting states impede visual processing or scene reasoning. It is a
form of scene graph reasoning, but based on executed tasks and
exploited object affordances.
The last two experiments, \textit{put spoon on plate then move to bowl} and 
\textit{place objects into empty containers} further require the
\textit{state keeper} to keep track of the state of the scene.  For
experimental purposes, a container is limited to containing one object
at a time. 
The third experiment asks the manipulator to move the spoon
(\textit{grasp}) onto the plate (\textit{contain}) in the first phase. 
At a later time, i.e., the second phase, the manipulator is then asked
to move the same spoon (\textit{grasp}) and put it into the bowl 
(\textit{contain}).  In this case, the task series are related.
The \textit{state keeper} retains awareness of the spoon's location
within the containers. 
The fourth experiment involves two \textit{grasp}-able objects and two 
\textit{contain}-able objects in a scene. It is an extension of the
first two experiments, except now there are two objects and two
containers. The task series consists of two tasks loosely related by
virtue of affordances.
In the first phase, one object is chosen to be placed into an empty
container with the container randomly chosen and specified. In the
second phase, the second object is required to be placed into an empty
container. The state memory ensures that the second object is placed
into the empty container.

\subsection{Methodology and Evaluation} \subseclabel{Evaluation}
This subsection describes the different metrics of a successful trial
for the five experiment classes.
For each experiment in a class, we conduct $10$ trials, and record the
outcome as a \textit{success} or a \textit{failure}. A final score
for each experiment is the number of successful trials for each
experiment. Details are provided below.

\subsubsection{Success in Each Scenario.}
The first three scenarios (seen categories, multiple objects, unseen categories) 
involve  simple movements with three common affordances, including \textit{grasp}, \textit{contain} and \textit{support}. Success for \textit{grasp} requires the manipulator to stably grasp the
target without dropping it. Success for \textit{contain} requires the
manipulator to put a small cuboid into a target container. 
Success for \textit{support} requires the manipulator to put the small
cuboid onto a target supportable surface without it falling off.
For each trial, the target is randomly placed (position and orientation)
in a visible and reachable area.

For the fourth scenario involving
task-oriented tasks, a successful \textit{pound peg into slot} trial
requires the manipulator to first grasp the target without dropping it,
and then fully pound the peg into the slot.
A successful \textit{cut through string} trial requires the manipulator
to first grasp the target tool without dropping it, and then fully cut
the string with the target. 

For the last experiment set involving the modified PDDL, a successful trial requires the execution result to
agree with the specified final state (one of the four commands discussed). For instance, if the predefined final state is
\textit{put knife into bowl}, success means the knife is inside the
bowl, which requires the manipulator to grasp the knife and place it
into the bowl. 

\begin{figure}[t]
         \centering
         \begin{tikzpicture}[inner sep=0pt,outer sep=0pt]
    
           \node (SF) at (0in,0in)
             {\includegraphics[height=1.20in]{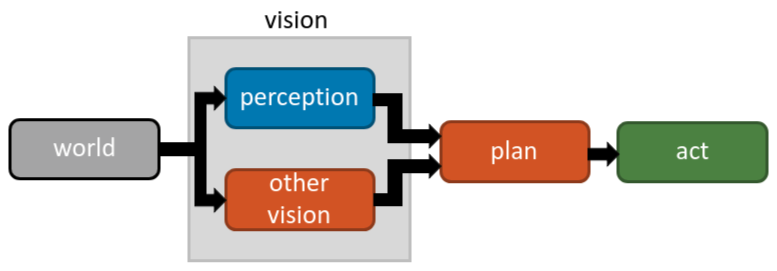}};
   
         \end{tikzpicture}
         \captionof{figure}{Illustration of the physical manipulation process. The vision includes the affordance detection (perception) and other vision processing functions. The visual results are then sent to the planning module followed by physical execution of our robotic manipulator (act module).}
            \label{perc_plan_act}
         \vspace*{-0.2in}   

\end{figure}

\subsubsection{Failure Categories.}
To properly evaluate and understand failure sources for the manipulation
experiments, failures are divided into one of the three consecutive
modules as shown in Fig. \ref{perc_plan_act}.
With reference to the block diagram, the vision sensor provides visual
input to the vision module. Within the vision module, two forms of
processing occur. The first is affordance segmentation, here denoted by
the \textit{Perception} block. The second is additional visual
processing to extract key geometric information for manipulation
planning (3D locations, regions, or $SE(3)$ poses). The extracted
signals are passed onto the planner (\textit{Plan} block), which uses them to determine the
world state and plan an action sequence to reach the target terminal
world state. The action sequence gets executed in open-loop, e.g, there
is no further visual processing to replan in the case of errors. The
manipulator has one shot to complete the task (\textit{Act} block).

Failures can occur in any of these modules. They are broken down into
three categories: \textit{perception}, \textit{planning}, and
\textit{action}.  Affordance related errors are counted towards the
\textit{perception} category.  To separate the affordance detection
error from general vision failures, the visual processing errors that
result in bad plans will be counted towards the \textit{planning}
category. 
Additionally, if the planning system generates a bad plan from good
information or fails to return a valid plan, the failure counts towards
\textit{planning}. 
Lastly, failures caused by bad physical movements leading to incomplete
tasks are recorded as \textit{action} failures.

\subsubsection{Common Failure Modes.}
To simplify the analysis of results, this section describes the failure
modes observed across all of the experiments. Several of the experiment
scenarios exhibited the same failure modes, thus it is best to describe
them in advance. For the affordance or \textit{perception} failure mode,
the error sources come in two types: direct and indirect. 
A direct error is when the affordance prediction is incorrect and fails
to identify the target affordance in the scene. Typically, the incorrect
affordance is assigned to the region and the target affordance does not
appear anywhere. A less common error is the failure to detect the object
as an object via the \textit{objectness} classifier, leading to no
affordance predictions. In both cases the affordance is not recognized
in the scene, and no planning nor manipulation actions can be taken.  
An indirect error is due to poor or noisy affordance segmentation. The
bad segmentation negatively impacts the subsequent geometric reasoning
in the other vision processes and leads to an incorrect plan. As a
consequence, the manipulator will fail to complete the task (be it
\textit{grasp}, \textit{contain}, \textit{support}, \textit{cut},
\textit{pound}, or \textit{scoop}). 

Moving to the vision side of the \textit{planning} errors, failures
include incorrect height or length estimation due to depth image noise
or bad relative geometry due to incorrect Aruco tag pose estimates.
The other source of \textit{planning} error occurs when the motion
planner fails to generate a feasible plan (it happened once).
The \textit{action} failure modes include dropping the object while
manipulator is moving, shifting of objects within the gripper while
being manipulated, or colliding with objects while moving. Though
collision could be a function of poor planning, it was common to have
this occur once or twice for an experiment indicating that execution
uncertainty or variance is the main factor. 
In-grasp shifts or drops cannot be corrected since there is no continual
perception processing to ensure nothing of significance changes during
open-loop execution.  Larger grasping forces and closed-loop execution
would improve on this failure type.

\begin{figure}[t]
  \centering
  \begin{tikzpicture}[inner sep=0pt,outer sep=0pt]
    \node (SF) at (0in,0in)
      {\includegraphics[height=1.5in,clip=true,trim=0in 0in 0in 0in]{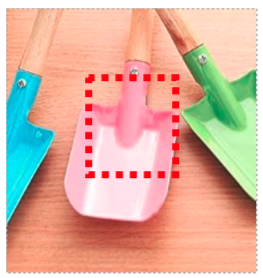}};
  \end{tikzpicture}
  \caption{Illustration of the non-flat surface of the shovel. The
  dashed red box indicated the non-flat part inside the supportable
  area.\label{shovel_explanation}}
\end{figure}

In several cases, the \textit{action} failure is a function of the
object geometry. Both the turners and the shovels do not lie flat on the
surface nor have trivial geometry (see Fig.~\ref{shovel_explanation}). 
Under normal circumstances a second arm would manipulate the object to
present a better relative geometry for placing the object within the
affordance action region. 
Or more tailored placement algorithms could be programmed for these
surfaces.  In the absence of a second arm or a custom place routine,
the robot simply attempts to place the object onto the surface. 
For the turners and the shovels, the object being placed (a cuboid)
sometimes slid or tumbled off of the surface. Since this
investigation does not consider the physics nor dynamics of the
manipulation actions, no modifications were made to correct for this
failure mode.  When describing these errors in the
experiments, the abbreviated description of sliding or tumbling from the
target object will be provided.

\begin{table*}[t]
  \centering
  \newcommand{\TSse}[1][1.9ex]{\rule{0pt}{#1}}
  \caption {Manipulation on Seen Categories\tablabel{physical_seen} }
  \small
  \addtolength{\tabcolsep}{1pt}
  \begin{tabular}{ | l || c | c | c ||  c | c | c || c |  c | c || c  | c | }
  \hline    
  & \multicolumn{3}{c||}{DeepGrasp}  
  & \multicolumn{3}{c||}{AffNet}  
  & \multicolumn{3}{c||}{\algName} 
  & \textit{similar} & \textit{affordance} \TSse[2.1ex]\\ \hline
  Object & Perc. & Plan & Act & Perc. & Plan & Act & Perc. & Plan & Act 
  & &\TSse[2.1ex]\\ \hline \hline
  knife           
    & 10 & 10 & 10    & 10 & 10 & 10    & 10 & 10 & 10
    & yes  &  grasp       
    \TSse \\ \hline
  spoon           
    & 10 & 10 & 10    & 10 & 10 & 10    & 10 & 10 & 10  
    & yes  &  grasp       
    \TSse \\ \hline
  mug             
    & \zr & \zr & \zr   & 10 & 10 & 10   & 10 & 10 & 9   
    & yes  &  contain     
    \TSse \\ \hline
  cup             
    & \zr & \zr & \zr   & 10 & 10 & 10   & 9 & 9 & 8   
    & yes  &  contain     
    \TSse \\ \hline 
  turner          
    & \zr & \zr & \zr   & 9 & 9 &  7   &  9 &  9 &  8   
    & yes  &  support     
    \TSse \\ \hline
  shovel          
    & \zr & \zr & \zr  & 8 & 8 &  6    & 8  & 8 & 7   
    & yes  &  support
    \TSse \\ \hline \hline
  knife           
    & 10 & 10 & 10   & 10 & 10  & 10    & 10 & 10  & 10
    & no   &  grasp       
    \TSse \\ \hline
  spoon           
    & 10 & 10 & 10    & 10 & 10 & 10    & 10 & 10 & 10  
    & no   &  grasp       
    \TSse \\ \hline
  mug             
    & \zr & \zr & \zr   & 10 &  10 & 10   & 10 & 10 & 10   
    & no   &  contain     
    \TSse \\ \hline
  cup             
    & \zr & \zr & \zr    & 10 &  10 & 10    & 10 & 10 & 10   
    & no   &  contain     
    \TSse \\ \hline 
  turner          
    & \zr & \zr & \zr    & 10 & 10 &  9     & 9 & 9 & 8   
    & no   &  support     
    \TSse \\ \hline
  shovel          
    & \zr & \zr & \zr   & 10 & 10 &  9    & 10 & 10 & 9   
    & no   &  support     
    \TSse \\ \hline \hline 
  grasp average         
    & \bf 10  & 10  & 10    & \bf 10  & 10  & 10    & \bf 10  & 10   & 10 
    & & \TSse[2.2ex]\\ \hline
  similar average          
    & \bf 3.3 & 3.3 & 3.3   & \bf 9.5 & 9.5 & 8.8   & \bf 9.3 & 9.3  & 8.7
    & & \TSse[2.2ex] \\ \hline
  all average         
    & \bf 3.3 & 3.3 & 3.3   & \bf 9.8 & 9.8 & 9.3   & \bf 9.6 & 9.6  & 9.1
    & & \TSse[2.2ex] \\ \hline
  \end{tabular}
  \addtolength{\tabcolsep}{-3pt}
\end{table*}

\section{Manipulation Results and Discussion} \label{result_sec}

\begin{table*}[t]
  \newcommand{\TSmo}[1][1.9ex]{\rule{0pt}{#1}}
  \centering
  \caption {Manipulation with Multiple Objects in Scene%
  \tablabel{physical_multiple}}
  \small
  \addtolength{\tabcolsep}{3pt}
  \begin{tabular}{ | l || c | c | c || c | c | c || c | c |  c || c  | }
  \hline    
  & \multicolumn{3}{c||}{DeepGrasp }  
  & \multicolumn{3}{c||}{AffNet }  
  & \multicolumn{3}{c||}{\algName}  
  & \textit{affordance} \TSmo[2.1ex]\\ \hline 
  Object & Perc. & Plan & Act & Perc. & Plan & Act & Perc. & Plan & Act &
  \\ \hline \hline
  knife    
    & 10 & 10 & 10      & 10 & 10 & 10    & 10 & 10 & 10     
    &  grasp       \\ \hline
  spoon    
    & 10 & 10 & 10      & 10 & 10 & 10    & 10 & 10 & 10     
    &  grasp       \\ \hline
  mug      
    & \zr & \zr & \zr   & 10 & 10 & 10    & 10 & 10 & 10     
    &  contain     \\ \hline 
  cup      
    & \zr & \zr & \zr   & 10 & 10 & 10    & 10 & 10 & 10     
    &  contain     \\ \hline
  turner   
    & \zr & \zr & \zr   & 10 & 10 & 8     & 10 & 10 & 7     
    &  support       \\ \hline
  shovel   
    & \zr & \zr & \zr   & 10 & 10 & 10    & 9  & 9  & 9     
    &  support       \\ \hline \hline 
  grasp average        
    & \bf 10  & 10 & 10   & \bf 10 &  10 &10 &  \bf 10 & 10 & 10 
    & \TSmo \\ \hline
  all average        
    & \bf 3.3 & 3.3 & 3.3   & \bf 10 & 10 & 9.7    & \bf 9.8 & 9.8 & 9.3
    & \TSmo \\ \hline
  \end{tabular}
  \addtolength{\tabcolsep}{-3pt}
\end{table*}

This section presents the experiment outcomes, discusses relative
performance properties, and attempts to synthesize the overall findings
associated to the experiments. Though the experiments attempt to
quantify how well the affordance predictions translate to task-based
manipulation, the tables will mark in bold the affordance
\textit{perception} outcomes for ease of identification.  Based on the
failure categories discussed earlier the \textit{planning} and
\textit{action} outcomes will also be quantified.  They will be in
standard font in the same rows and to the right of the bold font
\textit{perception} values.

\subsection{Affordance on Seen Categories\label{result_Seen_Categories}}
The proposed method is deployed and compared with state-of-the-art 
grasp \citep{chu2018deep} (\balg{DeepGrasp}) and 
affordance \citep{do2018affordancenet} (\balg{AffNet}) detectors, with
the results provided in Table \tabref{physical_seen}.
The \textit{grasp average} row indicates perfect processing and
execution for all three methods.  
The affordance recognition methods
successfully identified the graspable regions and used them to lift the
objects, thereby matching the performance of the specialized grasp
recognition implementation. 
Moving to the \textit{contain} affordance,
the \balg{AffNet} object-aware implementation had perfect visual
processing and execution. In contrast \balg{\algName} experienced a
single affordance recognition failure, and two execution failures. 
A similar trend in the performance loss occurs for the \textit{support}
affordance, whereby \balg{AffNet} had 3 affordance failures and 6
subsequent execution failures while \balg{\algName} had 4 affordance
failures and 4 subsequent execution failures. 
Affordance recognition for
\balg{\algName} demonstrates a less than 3\% performance drop over
\balg{AffNet} for affordance recognition and execution.
These outcomes indicate that the affordance segmentation performance
difference between the two methods, when applied to objects known by
\balg{AffNet}, does not influence task outcomes as much as the relative
F-measures would indicate. For the image-split \balg{AffNet} achieves a
0.80 weighted F-measure, while for the category-split \balg{\algName}
achieves a 0.69 weighted F-measure, which reflects a 13.8\% drop
relative to \balg{AffNet}.

For the \textit{perception} failures of \balg{AffNet}, 
2 were direct (shovel) 
and 
1 was indirect (turner).
For \textit{action} failures, all 
6 were due to sliding or tumbling (turner or shovel).
%
For \textit{perception} failures of \balg{\algName}, 
2 failures were direct (shovel)
and 
3 failures were indirect (cup and turner).
For \textit{action} failure, 2 failure were due to the end-effector
accidentally hitting the containers (cups and mugs) and 4 failures were
due to sliding or tumbling (turners and shovels).
The main difference lies in the indirect affordance perception failure
mode. 

\begin{table*}[t]
  \newcommand{\TSun}[1][1.9ex]{\rule{0pt}{#1}}
  \newcommand{\BSun}{}
  \centering
  \caption {Manipulation on Unseen Categories.\tablabel{physical_unseen}}
  \small
  \addtolength{\tabcolsep}{3pt}
  \begin{tabular}{ | l || c | c | c || c | c | c || c | c |  c || c | }
  \hline    
  & \multicolumn{3}{c||}{DeepGrasp }  
  & \multicolumn{3}{c||}{AffNet }  
  & \multicolumn{3}{c||}{\algName}  
  & \TSun[2.0ex] \\ \hline
  Object & Perc. & Plan & Act & Perc. & Plan & Act & Perc. & Plan & Act 
  & Affordance \TSun[2.0ex] \\ \hline \hline
  screwdriver    
    & 10 & 10 & 10     & \zr & \zr & \zr  & 10 & 10  & 10    
    &  grasp
    \TSun \\ \hline
  juice bottle   
    & 10 & 10  & 10     & \zr & \zr & \zr   & 10 & 10  & 10  
    &  grasp       
    \TSun \\ \hline
  mouse          
    & 10 & 10  & 10    & \zr & \zr & \zr  & 10  & 10  & 10     
    &  grasp     
    \TSun \\ \hline
  plate          
    & \zr & \zr & \zr   & \zr & \zr & \zr   & 10  & 10   & 10    
    &  contain     
    \TSun \\ \hline 
  jar            
    & \zr & \zr & \zr   & \zr & \zr & \zr   & 9 & 8 & 8      
    &  contain     
    \TSun \\ \hline
  can            
    & \zr & \zr & \zr   & \zr & \zr & \zr   & 10  & 10  & 10   
    &  contain     
    \TSun \\ \hline  
  griddle turner 
    & \zr & \zr & \zr   & \zr & \zr & \zr   & 10 & 10  &  8     
    &  support       
    \TSun \\ \hline
  grill spatula  
    & \zr & \zr & \zr   & \zr & \zr & \zr   & 10 & 10 &  9     
    &  support       
    \TSun \\ \hline
  pie server     
    & \zr & \zr & \zr   & \zr & \zr & \zr   &  9 &  9 &  9      
    &  support     
    \TSun \\ \hline \hline 
  grasp average        
    & \bf 10  & 10  & 10    & \bf 0 &  0 &  0  & \bf 10 & 10 & 10
    & \TSun[2.2ex] \\ \hline
  all average        
    & \bf 3.3 & 3.3 & 3.3   & \bf 0 &  0 &  0  & \bf 9.8 & 9.7& 9.3
    & \TSun[2.2ex] \\ \hline
  \end{tabular}
  \addtolength{\tabcolsep}{-2pt}
\end{table*}

\subsection{Affordance with Multiple Objects\label{result_Multi_Object}}
The grasp detector, \balg{DeepGrasp}, used for grasping comparisons is
designed to predict multiple grasps on multiple objects in a scene. 
On account of this design, grasp selection can be augmented to require
choosing a desired grasp from the candidate list (or subset thereof)
or to require specifying a region of interest from which the top
intersection grasp candidate must be chosen.  This experiment adopts the
latter modification (and thereby represents non-autonomous operation).  
The same affordances as in the previous experiment
are evaluated: \textit{grasp}, \textit{contain} and \textit{support}.
The experimental results are reported in Table \tabref{physical_multiple}\,.
%
For \textit{grasp}, all three methods achieve perfect perception,
planning, and execution.
%
Given that \balg{AffNet} was perfect across the other affordances, the
discussion looks at them in aggregate. 
There was a single indirect \textit{perception} failure for
\balg{\algName} for the shovel (\textit{support}).
%
Lastly \textit{AffNet} and \textit{\algName} had 2 and 3 execution
failures, respectively, for a differential of 1.
The 2 \textit{action} failures in \balg{AffNet} were due to
sliding/tumbling (turner). Likewise, the 3 \textit{action} failures in
\balg{\algName} were due to sliding/tumbling (turner).
%
%
Overall, \textit{\algName} shows a less than 2\% affordance perception
difference and a less than 2\% execution difference (for an overall less
than 4\% task completion difference).

\subsection{Affordance on Unseen Categories\label{result_Unseen_Categories}}
Given that \balg{AffNet} depends on the object prior and is limited to
the training categories, it will not be capable of recognizing
affordances for novel objects. Thus, even though it was run for this
scenario, the algorithm did not detect the presented objects and could
therefore not recognize any affordances.  In what follows \balg{AffNet}
will not be referenced since all tests failed. \balg{DeepGrasp} is not
restricted to specific object categories and can recognize graspable
regions of objects.  Table \tabref{physical_unseen} reports the outcomes
for this unseen categories experiment.  
Again, the \textit{grasp} affordance was perfectly perceived and executed 
for \balg{\algName}. The \textit{contain} and \textit{support}
affordances had 1 error each (out of 30 trials per affordance class)
for a 3\% affordance recognition error rate.  These errors were indirect
affordance perception errors.
When moving from the affordance perception to planning, then to action,
there was a total of 4 additional failures to complete the task.
The 1 \textit{planning} failure was due to incorrect height estimation (jar).
The 3 \textit{action} failures were sliding/tumbling errors (turner and
spatula).

Though there is a performance loss relative to \balg{AffNet} for seen
objects, the ability to operate without explicit object labels permits
more general operation of \balg{\algName}. In fact, affordance
prediction for these unseen objects matched \balg{AffNet} for the seen
objects.  The consistent task performance for these cases relative to
the earlier seen category tests (Tables \tabref{physical_seen} \&
\tabref{physical_multiple}), indicates good affordance generalization
performance for \balg{\algName} to unseen objects with known affordance
categories.

\subsection{Affordance for Task-Oriented Grasping\label{result_Task_Orient}}
These experiments move beyond \textit{pick-and-place} types of
affordance action tasks, which is what \textit{grasp}, \textit{contain}
and \textit{support} test. The affordances of \textit{pound} and
\textit{cut} require grasping an object and using it to achieve a given
goal for some other object in the world. In this case, the other object
has a known action region relative to an Aruco tag. Table \tabref{physical_taskoriented}
reports the results from these experiments, which involve seen and
unseen objects.

For the seen objects \balg{AffNet} and \balg{\algName} had the same
affordance perception performance, however \balg{\algName} had one more
planning failure than \balg{AffNet} (3 vs 2). The \textit{planning}
failures in all cases were due to visual processing errors (bad heights).
Compared to previous scenarios, the two designed task-oriented grasp
tasks in this scenario require good estimation of the heights of the peg
and the string.  Apart from the Aruco tag in the work space for estimating
relative transformation between camera and manipulator, another Aruco tag
is attached to the top surface of the box with the peg or string as
shown in Fig.  \ref{fig_task_oriented}. Due to the measurement
uncertainty from the depth image and the Aruco tags pose estimates, the
returned height estimates for the \textit{pound} or \textit{cut}
position are sometimes not accurate enough and led to misses.  The
\textit{action} failure for \balg{AffNet} was due to the hammer tilting
after being grasped. 

\begin{table*}[t]
  \centering
  \newcommand{\TSto}[1][1.9ex]{\rule{0pt}{#1}}
  \caption {Task-Oriented Grasping and Manipulation%
  \tablabel{physical_taskoriented}}
  \small
  \addtolength{\tabcolsep}{2pt}
  \begin{tabular}{ | l || c | c | c || c |  c | c || c |}
  \hline    
  & \multicolumn{3}{c||}{AffNet}  
  & \multicolumn{3}{c||}{\algName}  
  & \TSto[2ex] \\ \hline
  Object & Perc. & Plan & Act & Perc. & Plan & Act & Affordance 
  \TSto[2ex] \\ \hline \hline
  hammer        
    & 10  & 9   & 8     & 10  & 8   & 8   
    &  grasp, pound \TSto \\ \hline
  tenderizer    
    & \zr & \zr & \zr   & 9   & 8   & 8   
    &  grasp, pound \TSto \\ \hline
  knife         
    & 10  & 9   & 9     & 10  & 9   & 9   
    &  grasp, cut   \TSto \\ \hline 
  letter opener 
    & \zr & \zr & \zr   & 9   & 9   & 8   
    &  grasp, cut   \TSto \\ 
  \hline \hline
  seen average  
    & \bf 10  & 9.0 & 8.5   & \bf 10  & 8.5 & 8.5 
    & \TSto[2.2ex] \\ \hline
  all average       
    & \bf 5.0 & 4.5 & 4.3   & \bf 9.5 & 8.5 & 8.3 
    & \TSto[2.2ex] \\ \hline
  \end{tabular}
  \addtolength{\tabcolsep}{-2pt}
\end{table*}
\begin{table*}[t]
  \centering
  \newcommand{\TSsp}[1][1.9ex]{\rule{0pt}{#1}}
  \caption {Manipulation with Object Detector and State-PDDL.%
  \tablabel{physical_PDDL}}
  \small
  \begin{tabular}{ | l || c | c | c || c | c | c |}
  \hline    
  Activity & \multicolumn{3}{c||}{Success}  & \textit{Affordance} 
    & \textit{Detection} & \textit{State}\TSsp[2ex]\\ \hline
  & Perc. & Plan & Act & & & \TSsp[2ex]\\ \hline
  pick knife or spoon into bowl              
    & 10 & 10 & 10  
    & grasp, contain  &\ding{51} &             
    \TSsp \\ \hline
  select trowel or spoon	to scoop beans      
    & 10 & 10 & 8   
    &  grasp, scoop, contain & \ding{51} &     
    \TSsp \\ \hline
  grasp spoon to plate then move to bowl	    
    & 8 & 7 & 7      
    &  grasp, contain &  \ding{51} & \ding{51} 
    \TSsp \\ \hline 
  place objects into empty containers	    
    & 9 & 8 & 8       
    &  grasp, contain &  \ding{51} & \ding{51} 
    \TSsp[2.0ex] \\[-0.1ex] \hline \hline 
  average         
   & \bf 9.3 & 8.8 &  8.3  & & & \TSsp[2.2ex]\\ \hline
        \end{tabular}
\end{table*}

Moving to the unseen objects, the self-made tenderizer, see Fig.
\ref{fig_task_oriented}(c), case had an indirect \textit{perception}
failure for the \textit{grasp} affordance plus a \textit{planning}
failure (bad height).
For the letter opener (see Fig. \ref{fig_task_oriented}(d)), there was one
direct \textit{perception} failure for the \textit{grasp} affordance and
an \textit{action} failure. The letter opener tilted after the first
cut attempt so that the two additional attempts could not succeed.

The affordance recognition for \textit{\algName} in this set of
experiments is close to that of the earlier experiments but slightly
worse. There is a 5\% error rate at the affordance level with a further
12\% drop when translating to action.  This increased performance drop
from perception to action is a function of more perception-based
measurements outside of the affordance category. They accounted for 4 of
the 5 post-\textit{perception} module failures, or 80\%. In contrast the
earlier experiments have an aggregate \textit{planning} failure rate ten
times lower, at 8\%.
The current processing schemes are not robust to non-affordance
visual processing errors.  Nevertheless, these results demonstrate the
power of combining affordance reasoning with symbolic reasoning to plan
and execute manipulation activities.

\subsection{Affordance with Modified PDDL Pipeline\label{result_PDDL}}
Including state memory of prior actions to support contemporary vision
failures associated to overlapping or occluding objects supports
additional experiments that have goal state specification with
multiple, feasible solution plans. The experiments led to the outcomes reported in 
Table \tabref{physical_PDDL}.
%
For the first two experiments, the \textit{perception} component was
perfect with only \textit{action} failures affecting task completions.
The 2 \textit{action} failures were due to hitting the edge of the bowl
while performing the \textit{scoop} action primitive. 
%

The third and fourth experiments require the state memory component to
complete the specified task.  During successful operation for the third
experiment, the system correctly identifies the empty container with the
\textit{state keeper} as well as where the object of interest is placed on
the container (e.g, the \textit{plate}).
The 2 \textit{perception} failures for the third experiment,
\textit{grasp spoon to plate then move to bowl}, were direct and caused
by the \textit{objectness} of the plate not being detected by the
affordance detector.
The 1 \textit{action} failure was a motion planning failure.
%
For the fourth experiment, \textit{place objects into empty containers},
the system keeps track of the state of each container in order to accept
different combinations of the goal state. 
The 1 \textit{perception} failure was direct; the system did not perceive the
\textit{contain} affordance for the mug.  
The 1 \textit{planning} failure was due to the mug not being detected on
the object detector side (it is not an affordance related error).
The \textit{perception} error rate for this set of experiments
($\sim$7\%) is a bit higher than for the previous ones in aggregate
($\sim$3\%), which reflects the repeated use of the affordance
predictions as part of a sequential task. 
Meanwhile, the execution success rate drop of 10\% is consistent with
the previous task-oriented experiments. This experiment set demonstrates
that the \textit{\algName} perception module and PDDL planning module
provides the flexibility to specify and execute goal-oriented
manipulation tasks based on affordance informed state awareness of the
world.

\subsection{Discussion of Affordances and Manipulation}

Overall performance is around 97\% correct affordance recognition in
simple scenarios 
(seen categories, multiple objects, unseen categories), 
with a 92\% task completion success rate. 
The 3\% drop in affordance recognition could be improved through better
network design with regards to the overall detection and recognition
processes, while the subsequent 5\% performance drop could be improved
through better segmentation (more consistent regions) and improved
awareness of the scene geometry (e.g., orientation of support surfaces). 
Since the neural network model in the physical experiment is trained on
the UMD dataset, domain shift between training dataset and workspace may
be a source of performance loss. 
Though we showed that perfect estimation of the affordance map is not a
requirement for success in execution, better affordance segmentation
could be achieved by reducing the domain shift. A simple solution
involves finetuning on a small set of workspace or task-relevant
training data before deployment.     

In addition to simple manipulation tasks with affordance-based pick-and-place action demands, the task-oriented manipulation or sequential
manipulation experiments reflects moderately complex scenarios requiring
additional geometric processing to follow through on the task be
exploiting a part affordance. Affordance perception for these
demonstrated a 94\% success rate with \textit{\algName} when
aggregated, while execution led to a final success rate of 83\% (for an
11\% drop from perception to action).  The larger drop is mostly a
function of errors in the additional visual reasoning modules. The
\textit{perception} and \textit{action} outcomes are in agreement with
the simpler experiment given that these task-oriented experiments involve
sequential use of affordances to meet the task objective.

Aggregating all experiments, affordance recognition performance is 96\%
and task completion is 88\%.  Our earlier work has collected
grasping success rates for various deep learning algorithms (see
\cite{chu2018deep} and \cite{lin2020primshape}). 
For \textit{vision only} grasp detection, state-of-the-art success rates vary
from 87\% to 97\%. 
As a source of \textit{grasp} affordance information, the grasping
success rate across the experiments is at the upper end of this range
and matches the best methods.
Expanding further to include the additional affordance cases, the
success rates of the affordance perception module continues to lie at
the upper end of the vision-only success rates for published grasping
pipelines.
For \textit{embodied grasping} based on published state-of-the-art
grasping algorithms, success varies from 80\% to 97\% for the simple
case of grasping. 
The success rates of the simple affordance-based manipulation tasks lie
at the upper range of the embodied grasping range. Meanwhile, the
success rates of the various, more involved manipulation activities lie at the lower end of this range. 
However, the performance drop is a function of errors in other parts of the
perceive, plan, act pipeline. These outcomes can be improved through
better implementations of the supporting perception and planning
modules.  Across the experiments, 6 of the 7 affordances in the UMD
dataset were tested with the missing one being \textit{wrap-grasp}.
Considering the tasks involved, affordances employed, and the outcomes
achieved, \textit{\algName} exhibits state-of-the-art performance for
the affordance perception pipeline and successfully contributes towards
a perception to action pipeline.

In robotics, affordances have long been hypothesized to be a rich and
important source of action opportunities for seen objects. They can
contribute to symbolic planning needs through their integration into
a scene reasoning module. This paper designed a neural network that
successfully linked visually derived affordance recognition outputs to
physical manipulation plans. With affordance, robot agents with grasping
abilities may go beyond simple pick-and-place tasks and pursue
more general, goal-directed manipulations.
The proposed pipeline demonstrated both simple manipulations and
task-oriented manipulation activities.

\section{Conclusion}
This paper describes and evaluates an affordance-based scene
reasoning pipeline for robotic manipulation. The core component enabling
this is an object-agnostic affordance recognition neural network,
\balg{\algName}, for predicting the affordances of object parts within an image. 
The deep network learns to generalize affordance segmentation 
across unseen object categories in support of robotic manipulation. 
In particular, a self-attention mechanism within the affordance
segmentation branch selectively adapts to contextual dependencies within
each instance region.
Second, affordance category attributes guide the feature learning 
and compensate for the absence of object category priors. 
%
%
%
Evaluation of the \textit{perception} pipeline with the UMD dataset
used the novel category split for comparison to state-of-the-art methods,
including several image-based and region-based baselines.
Experiments with physical manipulation demonstrated the effectiveness
of the proposed framework for manipulating unseen object categories
in the real-world. It also demonstrated the successful integration of
affordance recognition with symbolic reasoning and planning in support
of goal-oriented manipulation. Future work aims to integrate 
affordance knowledge with scene graph knowledge to support more complex
manipulation sequences. It will also seek to plan activities involving
longer sequences of atomic actions.
All code and data will be publicly released.

\begin{funding}
This work was supported in part by NSF Award \#1605228.
\end{funding}


\bibliographystyle{SageH}
\bibliography{regular,crowdsourcing}

\begin{thebibliography}{90}
\providecommand{\natexlab}[1]{#1}
\providecommand{\url}[1]{\texttt{#1}}
\providecommand{\urlprefix}{URL }
\expandafter\ifx\csname urlstyle\endcsname\relax
  \providecommand{\doi}[1]{DOI:\discretionary{}{}{}#1}\else
  \providecommand{\doi}{DOI:\discretionary{}{}{}\begingroup
  \urlstyle{rm}\Url}\fi

\bibitem[{Akbari and Rosell(2015)}]{akbari2015ontological}
Akbari A and Rosell J (2015) Ontological physics-based motion planning for
  manipulation.
\newblock In: \emph{IEEE Conference on Emerging Technologies and Factory
  Automation}. pp. 1--7.

\bibitem[{Akbari and Rosell(2016)}]{akbari2016task}
Akbari A and Rosell J (2016) Task planning using physics-based heuristics on
  manipulation actions.
\newblock In: \emph{IEEE International Conference on Emerging Technologies and
  Factory Automation}. pp. 1--8.

\bibitem[{Aldoma et~al.(2012)Aldoma, Tombari and Vincze}]{aldoma2012supervised}
Aldoma A, Tombari F and Vincze M (2012) Supervised learning of hidden and
  non-hidden 0-order affordances and detection in real scenes.
\newblock In: \emph{IEEE International Conference on Robotics and Automation}.
  pp. 1732--1739.

\bibitem[{Antanas et~al.(2019)Antanas, Moreno, Neumann, de~Figueiredo,
  Kersting, Santos-Victor and De~Raedt}]{antanas2019semantic}
Antanas L, Moreno P, Neumann M, de~Figueiredo RP, Kersting K, Santos-Victor J
  and De~Raedt L (2019) Semantic and geometric reasoning for robotic grasping:
  a probabilistic logic approach.
\newblock \emph{Autonomous Robots} 43(6): 1393--1418.

\bibitem[{Bicchi and Kumar(2000)}]{bicchi2000robotic}
Bicchi A and Kumar V (2000) Robotic grasping and contact: A review.
\newblock In: \emph{IEEE International Conference on Robotics and Automation}.
  pp. 348--353.

\bibitem[{Bohg et~al.(2014)Bohg, Morales, Asfour and Kragic}]{bohg2014data}
Bohg J, Morales A, Asfour T and Kragic D (2014) Data-driven grasp synthesis—a
  survey.
\newblock \emph{IEEE Transactions on Robotics} 30(2): 289--309.

\bibitem[{Cambon et~al.(2004)Cambon, Gravot and Alami}]{cambon2004robot}
Cambon S, Gravot F and Alami R (2004) A robot task planner that merges symbolic
  and geometric reasoning.
\newblock In: \emph{European Conference on Artificial Intelligence}. pp.
  895--899.

\bibitem[{Chen et~al.(2013)Chen, Nugent and Okeyo}]{chen2013ontology}
Chen L, Nugent C and Okeyo G (2013) An ontology-based hybrid approach to
  activity modeling for smart homes.
\newblock \emph{IEEE Transactions on Human-Machine Systems} 44(1): 92--105.

\bibitem[{Chen et~al.(2018)Chen, Papandreou, Kokkinos, Murphy and
  Yuille}]{chen2018deeplab}
Chen LC, Papandreou G, Kokkinos I, Murphy K and Yuille AL (2018) Deeplab:
  Semantic image segmentation with deep convolutional nets, atrous convolution,
  and fully connected crfs.
\newblock \emph{IEEE Transactions on Pattern Analysis and Machine Intelligence}
  40(4): 834--848.

\bibitem[{Chen et~al.(2017)Chen, Papandreou, Schroff and
  Adam}]{chen2017rethinking}
Chen LC, Papandreou G, Schroff F and Adam H (2017) Rethinking atrous
  convolution for semantic image segmentation.
\newblock \emph{arXiv preprint arXiv:1706.05587} .

\bibitem[{Cheng et~al.(2013)Cheng, Sun, Griss, Davis, Li and
  You}]{cheng2013nuactiv}
Cheng HT, Sun FT, Griss M, Davis P, Li J and You D (2013) Nuactiv: Recognizing
  unseen new activities using semantic attribute-based learning.
\newblock In: \emph{International Conference on Mobile Systems, Applications,
  and Services}. ACM, pp. 361--374.

\bibitem[{Chu et~al.(2019{\natexlab{a}})Chu, Xu, Seguin and
  Vela}]{chu2019toward}
Chu FJ, Xu R, Seguin L and Vela PA (2019{\natexlab{a}}) Toward affordance
  detection and ranking on novel objects for real-world robotic manipulation.
\newblock \emph{IEEE Robotics and Automation Letters} 4(4): 4070--4077.

\bibitem[{Chu et~al.(2018)Chu, Xu and Vela}]{chu2018deep}
Chu FJ, Xu R and Vela PA (2018) Real-world multiobject, multigrasp detection.
\newblock \emph{IEEE Robotics and Automation Letters} 3(4): 3355--3362.

\bibitem[{Chu et~al.(2019{\natexlab{b}})Chu, Xu and Vela}]{chu2019learning}
Chu FJ, Xu R and Vela PA (2019{\natexlab{b}}) Learning affordance segmentation
  for real-world robotic manipulation via synthetic images.
\newblock \emph{IEEE Robotics and Automation Letters} 4(2): 1140--1147.

\bibitem[{Cordts et~al.(2016)Cordts, Omran, Ramos, Rehfeld, Enzweiler,
  Benenson, Franke, Roth and Schiele}]{Cordts2016Cityscapes}
Cordts M, Omran M, Ramos S, Rehfeld T, Enzweiler M, Benenson R, Franke U, Roth
  S and Schiele B (2016) The cityscapes dataset for semantic urban scene
  understanding.
\newblock In: \emph{IEEE Conference on Computer Vision and Pattern
  Recognition}. pp. 3213--3223.

\bibitem[{de~Silva et~al.(2013)de~Silva, Pandey, Gharbi and
  Alami}]{de2013towards}
de~Silva L, Pandey AK, Gharbi M and Alami R (2013) Towards combining htn
  planning and geometric task planning.
\newblock \emph{RSS Workshop on Combined Robot Motion Planning and AI Planning
  for Practical Applications} .

\bibitem[{Dehban et~al.(2016)Dehban, Jamone, Kampff and
  Santos-Victor}]{dehban2016denoising}
Dehban A, Jamone L, Kampff AR and Santos-Victor J (2016) Denoising
  auto-encoders for learning of objects and tools affordances in continuous
  space.
\newblock In: \emph{IEEE International Conference on Robotics and Automation}.
  pp. 4866--4871.

\bibitem[{Deng et~al.(2009)Deng, Dong, Socher, Li, Li and
  Fei-Fei}]{deng2009imagenet}
Deng J, Dong W, Socher R, Li LJ, Li K and Fei-Fei L (2009) Imagenet: A
  large-scale hierarchical image database.
\newblock In: \emph{IEEE Conference on Computer Vision and Pattern
  Recognition}. pp. 248--255.

\bibitem[{Ding et~al.(2018)Ding, Jiang, Shuai, Qun~Liu and
  Wang}]{ding2018context}
Ding H, Jiang X, Shuai B, Qun~Liu A and Wang G (2018) Context contrasted
  feature and gated multi-scale aggregation for scene segmentation.
\newblock In: \emph{IEEE Conference on Computer Vision and Pattern
  Recognition}. pp. 2393--2402.

\bibitem[{Do et~al.(2018)Do, Nguyen and Reid}]{do2018affordancenet}
Do TT, Nguyen A and Reid I (2018) Affordance{N}et: An end-to-end deep learning
  approach for object affordance detection.
\newblock In: \emph{IEEE International Conference on Robotics and Automation}.
  pp. 1--5.

\bibitem[{Duan et~al.(2012)Duan, Parikh, Crandall and
  Grauman}]{duan2012discovering}
Duan K, Parikh D, Crandall D and Grauman K (2012) Discovering localized
  attributes for fine-grained recognition.
\newblock In: \emph{IEEE Conference on Computer Vision and Pattern
  Recognition}. pp. 3474--3481.

\bibitem[{Erdem et~al.(2011)Erdem, Haspalamutgil, Palaz, Patoglu and
  Uras}]{erdem2011combining}
Erdem E, Haspalamutgil K, Palaz C, Patoglu V and Uras T (2011) Combining
  high-level causal reasoning with low-level geometric reasoning and motion
  planning for robotic manipulation.
\newblock In: \emph{IEEE International Conference on Robotics and Automation}.
  pp. 4575--4581.

\bibitem[{Farhadi et~al.(2009)Farhadi, Endres, Hoiem and
  Forsyth}]{farhadi2009describing}
Farhadi A, Endres I, Hoiem D and Forsyth D (2009) Describing objects by their
  attributes.
\newblock In: \emph{IEEE Conference on Computer Vision and Pattern
  Recognition}. pp. 1778--1785.

\bibitem[{Fikes and Nilsson(1971)}]{fikes1971strips}
Fikes RE and Nilsson NJ (1971) Strips: A new approach to the application of
  theorem proving to problem solving.
\newblock \emph{Artificial intelligence} 2(3-4): 189--208.

\bibitem[{Florence et~al.(2018)Florence, Manuelli and
  Tedrake}]{florence2018dense}
Florence PR, Manuelli L and Tedrake R (2018) Dense object nets: Learning dense
  visual object descriptors by and for robotic manipulation.
\newblock In: \emph{Conference on Robot Learning}. pp. 373--385.

\bibitem[{Fu et~al.(2019)Fu, Liu, Tian, Li, Bao, Fang and Lu}]{fu2019dual}
Fu J, Liu J, Tian H, Li Y, Bao Y, Fang Z and Lu H (2019) Dual attention network
  for scene segmentation.
\newblock In: \emph{IEEE Conference on Computer Vision and Pattern
  Recognition}. pp. 3146--3154.

\bibitem[{Gaschler et~al.(2015)Gaschler, Kessler, Petrick and
  Knoll}]{gaschler2015extending}
Gaschler A, Kessler I, Petrick RP and Knoll A (2015) Extending the knowledge of
  volumes approach to robot task planning with efficient geometric predicates.
\newblock In: \emph{IEEE International Conference on Robotics and Automation}.
  pp. 3061--3066.

\bibitem[{Ghallab et~al.(2004)Ghallab, Nau and Traverso}]{ghallab2004automated}
Ghallab M, Nau D and Traverso P (2004) \emph{Automated Planning: theory and
  practice}.
\newblock Elsevier.

\bibitem[{Gravot et~al.(2006)Gravot, Haneda, Okada and
  Inaba}]{gravot2006cooking}
Gravot F, Haneda A, Okada K and Inaba M (2006) Cooking for humanoid robot, a
  task that needs symbolic and geometric reasonings.
\newblock In: \emph{IEEE International Conference on Robotics and Automation}.
  pp. 462--467.

\bibitem[{Guo et~al.(2017)Guo, Sun, Liu, Kong, Fang and Xi}]{GuEtAl_ICRA2017}
Guo D, Sun F, Liu H, Kong T, Fang B and Xi N (2017) A hybrid deep architecture
  for robotic grasp detection.
\newblock In: \emph{IEEE International Conference on Robotics and Automation}.
  pp. 1609--1614.

\bibitem[{He et~al.(2017)He, Gkioxari, Doll{\'a}r and Girshick}]{he2017mask}
He K, Gkioxari G, Doll{\'a}r P and Girshick R (2017) Mask r-cnn.
\newblock In: \emph{IEEE International Conference on Computer Vision}. pp.
  2980--2988.

\bibitem[{Helmert(2006)}]{helmert2006fast}
Helmert M (2006) The fast downward planning system.
\newblock \emph{Journal of Artificial Intelligence Research} 26: 191--246.

\bibitem[{Jaderberg et~al.(2017)Jaderberg, Mnih, Czarnecki, Schaul, Leibo,
  Silver and Kavukcuoglu}]{jaderberg2016reinforcement}
Jaderberg M, Mnih V, Czarnecki WM, Schaul T, Leibo JZ, Silver D and Kavukcuoglu
  K (2017) Reinforcement learning with unsupervised auxiliary tasks.
\newblock In: \emph{International Conference on Learning Representations}.

\bibitem[{Kaelbling and Lozano-P{\'e}rez(2011)}]{kaelbling2011hierarchical}
Kaelbling LP and Lozano-P{\'e}rez T (2011) Hierarchical task and motion
  planning in the now.
\newblock In: \emph{IEEE International Conference on Robotics and Automation}.
  pp. 1470--1477.

\bibitem[{Kalashnikov et~al.(2018)Kalashnikov, Irpan, Pastor, Ibarz, Herzog,
  Jang, Quillen, Holly, Kalakrishnan, Vanhoucke and Levine}]{kalashnikov2018qt}
Kalashnikov D, Irpan A, Pastor P, Ibarz J, Herzog A, Jang E, Quillen D, Holly
  E, Kalakrishnan M, Vanhoucke V and Levine S (2018) {QT-Opt}: Scalable deep
  reinforcement learning for vision-based robotic manipulation.
\newblock In: \emph{Conference on Robot Learning}. pp. 651--673.

\bibitem[{Krasin et~al.(2016)Krasin, Duerig, Alldrin, Veit, Abu-El-Haija,
  Belongie, Cai, Feng, Ferrari, Gomes, Gupta, Narayanan, Sun, Chechik and
  Murphy}]{openimages}
Krasin I, Duerig T, Alldrin N, Veit A, Abu-El-Haija S, Belongie S, Cai D, Feng
  Z, Ferrari V, Gomes V, Gupta A, Narayanan D, Sun C, Chechik G and Murphy K
  (2016) Openimages: A public dataset for large-scale multi-label and
  multi-class image classification.
\newblock \emph{Dataset available from https://github.com/openimages} .

\bibitem[{Krizhevsky(2009)}]{krizhevsky2009learning}
Krizhevsky A (2009) Learning multiple layers of features from tiny images.
\newblock Technical report, Citeseer.

\bibitem[{Krizhevsky et~al.(2012)Krizhevsky, Sutskever and
  Hinton}]{krizhevsky2012imagenet}
Krizhevsky A, Sutskever I and Hinton GE (2012) Imagenet classification with
  deep convolutional neural networks.
\newblock In: \emph{Advances in Neural Information Processing Systems}. pp.
  1097--1105.

\bibitem[{Kumar et~al.(2011)Kumar, Berg, Belhumeur and
  Nayar}]{kumar2011describable}
Kumar N, Berg A, Belhumeur PN and Nayar S (2011) Describable visual attributes
  for face verification and image search.
\newblock \emph{IEEE Transactions on Pattern Analysis and Machine Intelligence}
  33(10): 1962--1977.

\bibitem[{Kumar et~al.(2009)Kumar, Berg, Belhumeur and
  Nayar}]{kumar2009attribute}
Kumar N, Berg AC, Belhumeur PN and Nayar SK (2009) Attribute and simile
  classifiers for face verification.
\newblock In: \emph{IEEE International Conference on Computer Vision}. pp.
  365--372.

\bibitem[{Lab(2013)}]{cornell2013}
Lab RL (2013) Cornell grasping dataset.
\newblock \url{http://pr.cs.cornell.edu/grasping/rect_data/data.php}.
\newblock Accessed: 2019-09-01.

\bibitem[{Lenz et~al.(2015)Lenz, Lee and Saxena}]{lenz2015deep}
Lenz I, Lee H and Saxena A (2015) Deep learning for detecting robotic grasps.
\newblock \emph{The International Journal of Robotics Research} 34(4-5):
  705--724.

\bibitem[{Levihn and Stilman(2014)}]{levihn2014using}
Levihn M and Stilman M (2014) Using environment objects as tools:
  Unconventional door opening.
\newblock In: \emph{IEEE/RSJ International Conference on Intelligent Robots and
  Systems}. pp. 2502--2508.

\bibitem[{Levine et~al.(2016)Levine, Pastor, Krizhevsky and
  Quillen}]{levine2016learning}
Levine S, Pastor P, Krizhevsky A and Quillen D (2016) Learning hand-eye
  coordination for robotic grasping with large-scale data collection.
\newblock In: \emph{International Symposium on Experimental Robotics}. pp.
  173--184.

\bibitem[{Li et~al.(2010)Li, Su, Lim and Fei-Fei}]{li2010objects}
Li LJ, Su H, Lim Y and Fei-Fei L (2010) Objects as attributes for scene
  classification.
\newblock In: \emph{European Conference on Computer Vision}. Springer, pp.
  57--69.

\bibitem[{Li(2017)}]{li2017deep}
Li Y (2017) Deep reinforcement learning: An overview.
\newblock \emph{arXiv preprint arXiv:1701.07274} .

\bibitem[{Lin et~al.(2020)Lin, Tang, Chu and Vela}]{lin2020primshape}
Lin Y, Tang C, Chu FJ and Vela PA (2020)
  \href{https://arxiv.org/pdf/1909.08508.pdf}{Using Synthetic Data and Deep
  Networks to Recognize Primitive Shapes for Object Grasping}.
\newblock \emph{IEEE International Conference on Robotics and Automation} .

\bibitem[{Liu et~al.(2020)Liu, Chen, Wurm and von Wichert}]{liu2020table}
Liu Z, Chen D, Wurm KM and von Wichert G (2020) Table-top scene analysis using
  knowledge-supervised mcmc.
\newblock \emph{arXiv preprint arXiv:2002.08417} .

\bibitem[{Liu et~al.(2016)Liu, Luo, Qiu, Wang and Tang}]{liu2016deepfashion}
Liu Z, Luo P, Qiu S, Wang X and Tang X (2016) Deepfashion: Powering robust
  clothes recognition and retrieval with rich annotations.
\newblock In: \emph{IEEE Conference on Computer Vision and Pattern
  Recognition}. pp. 1096--1104.

\bibitem[{Margolin et~al.(2014)Margolin, Zelnik-Manor and
  Tal}]{margolin2014evaluate}
Margolin R, Zelnik-Manor L and Tal A (2014) How to evaluate foreground maps?
\newblock In: \emph{IEEE Conference on Computer Vision and Pattern
  Recognition}. pp. 248--255.

\bibitem[{Marthi et~al.(2007)Marthi, Russell and Wolfe}]{marthi2007angelic}
Marthi B, Russell SJ and Wolfe JA (2007) Angelic semantics for high-level
  actions.
\newblock In: \emph{International Conference on Automated Planning and
  Scheduling}. pp. 232--239.

\bibitem[{McDermott et~al.(1998)McDermott, Ghallab, Howe, Knoblock, Ram,
  Veloso, Weld and Wilkins}]{mcdermott1998pddl}
McDermott D, Ghallab M, Howe A, Knoblock C, Ram A, Veloso M, Weld D and Wilkins
  D (1998) {PDDL}-the planning domain definition language .

\bibitem[{Migimatsu and Bohg(2020)}]{migimatsu2020object}
Migimatsu T and Bohg J (2020) Object-centric task and motion planning in
  dynamic environments.
\newblock \emph{IEEE Robotics and Automation Letters} 5(2): 844--851.

\bibitem[{Myers et~al.(2015)Myers, Teo, Ferm{\"u}ller and
  Aloimonos}]{myers2015affordance}
Myers A, Teo CL, Ferm{\"u}ller C and Aloimonos Y (2015) Affordance detection of
  tool parts from geometric features.
\newblock In: \emph{IEEE International Conference on Robotics and Automation}.
  pp. 1374--1381.

\bibitem[{Ngiam et~al.(2011)Ngiam, Khosla, Kim, Nam, Lee and
  Ng}]{ngiam2011multimodal}
Ngiam J, Khosla A, Kim M, Nam J, Lee H and Ng AY (2011) Multimodal deep
  learning.
\newblock In: \emph{International Conference on Machine Learning}. ACM, pp.
  689--696.

\bibitem[{Nguyen et~al.(2016{\natexlab{a}})Nguyen, Kanoulas, Caldwell and
  Tsagarakis}]{nguyen2016detecting}
Nguyen A, Kanoulas D, Caldwell DG and Tsagarakis NG (2016{\natexlab{a}})
  Detecting object affordances with convolutional neural networks.
\newblock In: \emph{IEEE/RSJ International Conference on Intelligent Robots and
  Systems}. pp. 2765--2770.

\bibitem[{Nguyen et~al.(2016{\natexlab{b}})Nguyen, Kanoulas, Caldwell and
  Tsagarakis}]{nguyen2016preparatory}
Nguyen A, Kanoulas D, Caldwell DG and Tsagarakis NG (2016{\natexlab{b}})
  Preparatory object reorientation for task-oriented grasping.
\newblock In: \emph{IEEE/RSJ International Conference on Intelligent Robots and
  Systems}. pp. 893--899.

\bibitem[{Nguyen et~al.(2017)Nguyen, Kanoulas, Caldwell and
  Tsagarakis}]{nguyen2017object}
Nguyen A, Kanoulas D, Caldwell DG and Tsagarakis NG (2017) Object-based
  affordances detection with convolutional neural networks and dense
  conditional random fields.
\newblock In: \emph{IEEE/RSJ International Conference on Intelligent Robots and
  Systems}. pp. 5908--5915.

\bibitem[{Pandey et~al.(2012)Pandey, Saut, Sidobre and
  Alami}]{pandey2012towards}
Pandey AK, Saut JP, Sidobre D and Alami R (2012) Towards planning human-robot
  interactive manipulation tasks: Task dependent and human oriented autonomous
  selection of grasp and placement.
\newblock In: \emph{IEEE RAS and EMBS International Conference on Biomedical
  Robotics and Biomechatronics}. pp. 1371--1376.

\bibitem[{Pednault(1994)}]{pednault1994adl}
Pednault EP (1994) Adl and the state-transition model of action.
\newblock \emph{Journal of logic and computation} 4(5): 467--512.

\bibitem[{Plaku and Hager(2010)}]{plaku2010sampling}
Plaku E and Hager GD (2010) Sampling-based motion and symbolic action planning
  with geometric and differential constraints.
\newblock In: \emph{IEEE International Conference on Robotics and Automation}.
  pp. 5002--5008.

\bibitem[{Redmon and Angelova(2015)}]{redmon2015real}
Redmon J and Angelova A (2015) Real-time grasp detection using convolutional
  neural networks.
\newblock In: \emph{IEEE International Conference on Robotics and Automation}.
  pp. 1316--1322.

\bibitem[{Ren et~al.(2015)Ren, He, Girshick and Sun}]{ren2015faster}
Ren S, He K, Girshick R and Sun J (2015) Faster {R-CNN}: Towards real-time
  object detection with region proposal networks.
\newblock In: \emph{Advances in Neural Information Processing Systems}. pp.
  91--99.

\bibitem[{Rezapour~Lakani et~al.(2019)Rezapour~Lakani,
  Rodr{\'i}guez-S{\'a}nchez and Piater}]{Rezapour_Lakani2018}
Rezapour~Lakani S, Rodr{\'i}guez-S{\'a}nchez AJ and Piater J (2019) Towards
  affordance detection for robot manipulation using affordance for parts and
  parts for affordance.
\newblock \emph{Autonomous Robots} 43(5): 1155--1172.

\bibitem[{Richter et~al.(2016)Richter, Vineet, Roth and
  Koltun}]{Richter_2016_ECCV}
Richter SR, Vineet V, Roth S and Koltun V (2016) Playing for data: {G}round
  truth from computer games.
\newblock In: \emph{European Conference on Computer Vision}. Springer, pp.
  102--118.

\bibitem[{Ros et~al.(2016)Ros, Sellart, Materzynska, Vazquez and
  Lopez}]{ros2016synthia}
Ros G, Sellart L, Materzynska J, Vazquez D and Lopez AM (2016) The synthia
  dataset: A large collection of synthetic images for semantic segmentation of
  urban scenes.
\newblock In: \emph{IEEE Conference on Computer Vision and Pattern
  Recognition}. pp. 3234--3243.

\bibitem[{Roy and Todorovic(2016)}]{roy2016multi}
Roy A and Todorovic S (2016) A multi-scale cnn for affordance segmentation in
  rgb images.
\newblock In: \emph{European Conference on Computer Vision}. Springer, pp.
  186--201.

\bibitem[{Sawatzky and Gall(2017)}]{sawatzky2017adaptive}
Sawatzky J and Gall J (2017) Adaptive binarization for weakly supervised
  affordance segmentation.
\newblock In: \emph{IEEE International Conference on Computer Vision
  Workshops}. pp. 1383--1391.

\bibitem[{Sawatzky et~al.(2017)Sawatzky, Srikantha and
  Gall}]{sawatzky2017weakly}
Sawatzky J, Srikantha A and Gall J (2017) Weakly supervised affordance
  detection.
\newblock In: \emph{IEEE Conference on Computer Vision and Pattern
  Recognition}. pp. 5197--5206.

\bibitem[{Saxena et~al.(2008)Saxena, Driemeyer and Ng}]{saxena2008robotic}
Saxena A, Driemeyer J and Ng AY (2008) Robotic grasping of novel objects using
  vision.
\newblock \emph{The International Journal of Robotics Research} 27(2):
  157--173.

\bibitem[{Shimoga(1996)}]{shimoga1996robot}
Shimoga KB (1996) Robot grasp synthesis algorithms: A survey.
\newblock \emph{The International Journal of Robotics Research} 15(3):
  230--266.

\bibitem[{Simonyan and Zisserman(2015)}]{simonyan2014very}
Simonyan K and Zisserman A (2015) Very deep convolutional networks for
  large-scale image recognition.
\newblock In: \emph{International Conference on Learning Representations}.

\bibitem[{Srikantha and Gall(2016)}]{srikantha2016weakly}
Srikantha A and Gall J (2016) Weakly supervised learning of affordances.
\newblock \emph{arXiv preprint arXiv:1605.02964} .

\bibitem[{Srivastava et~al.(2014)Srivastava, Fang, Riano, Chitnis, Russell and
  Abbeel}]{srivastava2014combined}
Srivastava S, Fang E, Riano L, Chitnis R, Russell S and Abbeel P (2014)
  Combined task and motion planning through an extensible planner-independent
  interface layer.
\newblock In: \emph{IEEE International Conference on Robotics and Automation}.
  pp. 639--646.

\bibitem[{Stilman and Kuffner(2005)}]{stilman2005navigation}
Stilman M and Kuffner JJ (2005) Navigation among movable obstacles: Real-time
  reasoning in complex environments.
\newblock \emph{International Journal of Humanoid Robotics} 2(04): 479--503.

\bibitem[{Su{\'a}rez-Hern{\'a}ndez et~al.(2018)Su{\'a}rez-Hern{\'a}ndez,
  Aleny{\`a} and Torras}]{suarez2018interleaving}
Su{\'a}rez-Hern{\'a}ndez A, Aleny{\`a} G and Torras C (2018) Interleaving
  hierarchical task planning and motion constraint testing for dual-arm
  manipulation.
\newblock In: \emph{IEEE/RSJ International Conference on Intelligent Robots and
  Systems}. pp. 4061--4066.

\bibitem[{Sui et~al.(2017)Sui, Xiang, Jenkins and Desingh}]{sui2017goal}
Sui Z, Xiang L, Jenkins OC and Desingh K (2017) Goal-directed robot
  manipulation through axiomatic scene estimation.
\newblock \emph{The International Journal of Robotics Research} 36(1): 86--104.

\bibitem[{Sun et~al.(2013)Sun, Bo and Fox}]{sun2013attribute}
Sun Y, Bo L and Fox D (2013) Attribute based object identification.
\newblock In: \emph{IEEE International Conference on Robotics and Automation}.
  pp. 2096--2103.

\bibitem[{Tenorth and Beetz(2009)}]{tenorth2009knowrob}
Tenorth M and Beetz M (2009) Knowrob—knowledge processing for autonomous
  personal robots.
\newblock In: \emph{IEEE/RSJ international Conference on Intelligent Robots and
  Systems}. pp. 4261--4266.

\bibitem[{Toussaint et~al.(2010)Toussaint, Plath, Lang and
  Jetchev}]{toussaint2010integrated}
Toussaint M, Plath N, Lang T and Jetchev N (2010) Integrated motor control,
  planning, grasping and high-level reasoning in a blocks world using
  probabilistic inference.
\newblock In: \emph{IEEE International Conference on Robotics and Automation}.
  pp. 385--391.

\bibitem[{Ugur and Piater(2015)}]{ugur2015bottom}
Ugur E and Piater J (2015) Bottom-up learning of object categories, action
  effects and logical rules: From continuous manipulative exploration to
  symbolic planning.
\newblock In: \emph{IEEE International Conference on Robotics and Automation}.
  pp. 2627--2633.

\bibitem[{Vaswani et~al.(2017)Vaswani, Shazeer, Parmar, Uszkoreit, Jones,
  Gomez, Kaiser and Polosukhin}]{vaswani2017attention}
Vaswani A, Shazeer N, Parmar N, Uszkoreit J, Jones L, Gomez AN, Kaiser {\L} and
  Polosukhin I (2017) Attention is all you need.
\newblock In: \emph{Advances in Neural Information Processing Systems}. pp.
  5998--6008.

\bibitem[{Winkler et~al.(2012)Winkler, Bartels, M{\"o}senlechner and
  Beetz}]{winkler2012knowledge}
Winkler J, Bartels G, M{\"o}senlechner L and Beetz M (2012) Knowledge enabled
  high-level task abstraction and execution.
\newblock In: \emph{Conference on Advances in Cognitive Systems}, volume~2.
  Citeseer, pp. 131--148.

\bibitem[{Yang et~al.(2015)Yang, Li, Fermuller and Aloimonos}]{yang2015robot}
Yang Y, Li Y, Fermuller C and Aloimonos Y (2015) Robot learning manipulation
  action plans by" watching" unconstrained videos from the world wide web.
\newblock In: \emph{AAAI Conference on Artificial Intelligence}.

\bibitem[{Yuan and Wang(2018)}]{yuan2018ocnet}
Yuan Y and Wang J (2018) Ocnet: Object context network for scene parsing.
\newblock \emph{arXiv preprint arXiv:1809.00916} .

\bibitem[{Zeng et~al.(2018{\natexlab{a}})Zeng, Song, Welker, Lee, Rodriguez and
  Funkhouser}]{zeng2018learning}
Zeng A, Song S, Welker S, Lee J, Rodriguez A and Funkhouser T
  (2018{\natexlab{a}}) Learning synergies between pushing and grasping with
  self-supervised deep reinforcement learning.
\newblock In: \emph{IEEE/RSJ International Conference on Intelligent Robots and
  Systems}. pp. 4238--4245.

\bibitem[{Zeng et~al.(2018{\natexlab{b}})Zeng, Zhou, Sui and
  Jenkins}]{zeng2018semantic}
Zeng Z, Zhou Z, Sui Z and Jenkins OC (2018{\natexlab{b}}) Semantic robot
  programming for goal-directed manipulation in cluttered scenes.
\newblock In: \emph{IEEE International Conference on Robotics and Automation}.
  pp. 7462--7469.

\bibitem[{Zhang et~al.(2018)Zhang, Dana, Shi, Zhang, Wang, Tyagi and
  Agrawal}]{zhang2018context}
Zhang H, Dana K, Shi J, Zhang Z, Wang X, Tyagi A and Agrawal A (2018) Context
  encoding for semantic segmentation.
\newblock In: \emph{IEEE Conference on Computer Vision and Pattern
  Recognition}. pp. 7151--7160.

\bibitem[{Zhang et~al.(2015)Zhang, Luo, Loy and Tang}]{zhang2015learning}
Zhang Z, Luo P, Loy CC and Tang X (2015) Learning deep representation for face
  alignment with auxiliary attributes.
\newblock \emph{IEEE Transactions on Pattern Analysis and Machine Intelligence}
  38(5): 918--930.

\bibitem[{Zhao et~al.(2017)Zhao, Shi, Qi, Wang and Jia}]{zhao2017pyramid}
Zhao H, Shi J, Qi X, Wang X and Jia J (2017) Pyramid scene parsing network.
\newblock In: \emph{IEEE Conference on Computer Vision and Pattern
  Recognition}. pp. 2881--2890.

\end{thebibliography}

\end{document}